\documentclass[runningheads]{llncs}

\usepackage{accv}

\usepackage{accvabbrv}

\usepackage{graphicx}
\usepackage{booktabs}

\usepackage[accsupp]{axessibility}  
\usepackage{multirow, tabularx}

\newcommand{\SErrMin}{{$\textup{SErr-S}$\;}}
\newcommand{\mb}[1]{\mathbf{#1}}
\definecolor{goldenrod}{RGB}{210,160,20}

\usepackage{algorithm}
\usepackage{algpseudocode}

\usepackage{hyperref}

\usepackage{orcidlink}

\begin{document}


\title{GeoBlur: Epipolar Geometry Estimation\\ from a Single Motion-Blurred Image}


\author{Bao-Long Tran \and
Cuong Le  \and
Fredrik Viksten  \and 
Per-Erik Forss{\'e}n }

\titlerunning{GeoBlur}

\authorrunning{B.L. Tran et al.}

\institute{Link{\"o}ping University, Sweden \\
\email{\{firstname.lastname\}@liu.se}}

\maketitle

\begin{abstract}
Relative camera pose geometry, formulated via fundamental matrix estimation, is a challenging problem in many robotics and VR/AR applications.
These applications occasionally contain fast monocular camera motion, which severely blurs the image and prevents the use of traditional multi-view geometry methods for camera pose estimation.
To handle these cases, we propose GeoBlur, a framework for estimating the fundamental matrix and recovering relative camera pose directly from a single motion-blurred image, using the motion cues from blur artifacts.
GeoBlur first predicts the visual correspondences between two time instances {\it within the camera exposure window};
then, it infers the fundamental matrix by solving the single-frame epipolar geometry problem under time-direction ambiguity. 
The resulting fundamental matrix is unique up to transposition, reflecting the inherent ambiguity in the direction of time.
GeoBlur improves performance on synthetic and hybrid benchmarks while remaining competitive with prior work on real motion-blur data.
We further demonstrate the use of GeoBlur on the downstream task on single frame motion segmentation.
The code is available \href{https://github.com/Bao-Long-Tran/GeoBlur}{\textcolor{blue}{here}}.


\end{abstract}

\section{Introduction}
\label{sec:intro}
The fundamental matrix encodes the relative pose between two uncalibrated cameras, in the form of the underlying epipolar geometry~\cite{hartley_zisserman2004}.
The matrix is commonly used for 3D point triangulation, and for verifying the geometric consistency of correspondences across two views.
It is a key component in many computer vision applications, \eg visual odometry (VO) \cite{nister2004visual,cvivsic2022soft2}, visual localization~\cite{brachmann18}, structure from motion (SfM)~\cite{schonberger2016structure,pan2024glomap}, and practical applications such as local-motion object segmentation \cite{argaw2021optical,Jain_2017_CVPR}. 
To date, the importance of the fundamental matrix has been recognized primarily in two-view settings.
Well-known approaches for fundamental matrix estimation, \eg the 7-point or the 8-point algorithms~\cite{hartley_zisserman2004}, require 2D correspondences between separate views, limiting their applicability to two-view/multi-view camera setups~\cite{faugeras1992can,hartley_zisserman2004,ranftl2018deep,poursaeed2018deep,ding2024fundamental}.
\begin{figure}[t]
    \centering
    \includegraphics[width=0.99\textwidth]{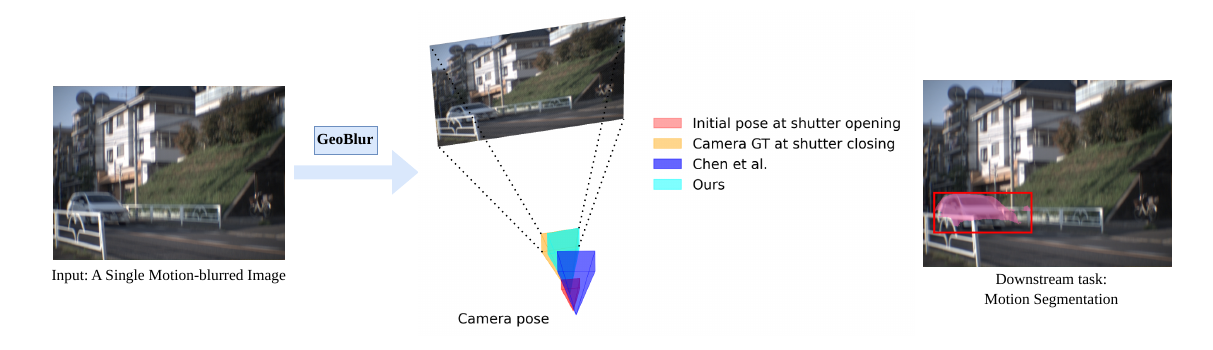}
    \caption{
    From a motion-blurred image (left), GeoBlur estimates the motion geometry that benefits the inference of relative camera pose during image exposure (middle), and motion segmentation (right).
    The (\textcolor{red}{red}) cone is the initial camera at shutter opening, and (\textcolor{goldenrod}{yellow}) cone corresponds to the ground-truth (GT) camera pose at shutter closing.
    GeoBlur captures the pose at shutter closing (\textcolor{cyan}{cyan}) with higher accuracy than the prior work by Chen et al.~\cite{chen2025imageimu}. This example is from scene 46 of the RBI dataset\cite{Zhong_2023_CVPR}.
    }
    \label{fig:teaser}
\end{figure}
%
On the other hand, many robotics and AR/VR applications only adopt the single-camera setup~\cite{Davison2007MonoSLAM,MurArtal2015MonocularVR}, which makes motion perception from single blurry frames as relevant problem~\cite{chen2025imageimu}.
These applications occasionally contain fast camera motion, which causes blurry patterns in the captured footage during exposure time.
Prior work treats these blurry patterns as undesired artifacts or motion noises; and usually learns to de-noise them~\cite{fang2023self,kohler2012recording}.
However, these large motion-blur paths can provide important cues about the underlying camera motion, yet remain underexplored in the literature.
Instead of treating them as degradations, we exploit the latent cues and geometric constraints from these blur artifacts to infer the motion geometry during the exposure window, expressed in the form of a fundamental matrix.
The relative camera pose can be derived from the fundamental matrix and is predominantly conducted in two-view settings.
A prior work~\cite{chen2025imageimu} extends this classical perspective to the intra-exposure setting, where only one blurred image is available.
While deblurring methods that use pixel-wise blur kernels can provide an approximation of local pixel-wise camera rotation, or in-plane translation, they do not capture the global camera motion~\cite{fergus2006removing,whyte2012non,fang2023self}.
In contrast, our method recovers the global camera motion by robustly estimating the fundamental matrix from a single motion-blurred input.
\cref{fig:teaser} shows an accurate pose estimation example from a single blurry image using our method.

The smear paths are formed by the projection of 3D points onto the image plane over the exposure time; therefore, pixels along the same smear path can be regarded as correspondences. 
However, the temporal direction of each correspondence, which is required to formulate the conventional epipolar constraint, is unknown.
This makes multi-view methods inapplicable, \eg the 8-point algorithm and reprojection-error optimization~\cite{hartley_zisserman2004}.
To address the challenges, we propose GeoBlur (short for \emph{Geometry from Motion Blur}), a novel framework for estimating the fundamental matrix encoding the relative camera pose between shutter opening and closing. 
Our method consists of two stages: 1) smear path estimation from blurred images, followed by 2) robust fundamental matrix estimation from the predicted smear paths.
First, a network learns to predict per-pixel smear paths, together with an uncertainty measure.
Here we propose to use the double-angle representation~\cite{bigun87} to describe a smear path.
Second, the fundamental matrix is estimated from the two endpoints of the predicted smear paths using a RANSAC-based procedure to improve robustness against local motion.
Note that, the exact smear path shape is not critical in our setting, as the epipolar constraint depends only on the two end points.
We describe how we address the time-direction ambiguity of each correspondence in~\cref{subsec:epipolar-est}.
Our method further enables the application of motion segmentation from a single blurred image.

To assess our work, we compare our method with both \emph{direct} and \emph{indirect} methods. 
Direct methods refer to approaches that estimate the relative pose, or an equivalent geometric representation, directly from a single blurred image. 
These include our designed network that directly regresses the nine entries of the fundamental matrix, inspired by direct regression study in~\cite{ranftl2018deep} in two-view setting; as well as the approach by Chen et al.~\cite{chen2025imageimu}, which estimates instantaneous camera velocity and can be used to derive the relative camera pose over the exposure time.
The indirect baseline refers to a pipeline that interpolates the first and last sharp images from a blurred input, inspired by the work in~\cite{Zhong_2023_CVPR}, then treats them as a two-view pair.
It generates instantaneous sharp frames using a pre-trained model~\cite{Zhong_2023_CVPR}, and then applies conventional methods to the first and last frames.
Experiments are conducted on the real-world RBI dataset~\cite{Zhong_2023_CVPR}, a hybrid-realistic motion-blur dataset based on GoPro~\cite{nah2017deep}, and synthetic datasets derived from Monkaa and Driving~\cite{mayer2016large}.
We further demonstrate the practical applications of our work via a downstream task of local motion segmentation, in which we use the predicted smear paths and the fundamental matrix to identify locally moving objects in blurry images.
Our contributions are summarized as follows:
\begin{itemize}
    \item We propose GeoBlur, a framework that robustly estimates the fundamental matrix from a single motion-blurred image. 
    \item We show that the double-angle formulation improves smear path learning, and that smear-path uncertainty allows robust geometry estimation.
    \item We demonstrate the applications of our method on the downstream task of camera pose estimation within the exposure time and motion segmentation.
\end{itemize}
\vspace{-15pt}
\section{Related Work}
\label{sec:related_work}

\subsection{Motion Blur}
\label{subsec:related_blur}

The estimation of pixel-wise motion from consecutive images (\ie optical flow) has received significant attention~\cite{sun2018pwc,dosovitskiy2015flownet}.
In contrast, pixel-wise motion estimation from one blurry image, represented by motion-blur smear paths, remains highly challenging due to the similar appearance of smear paths and other structures in the image~\cite{rekleitis1996optical,argaw2021optical}.
The estimation of the correlation between the smear paths and their cause, motion, was first studied by~\cite{chen1996image}. 
While the point-spread function (PSF), which describes blur characteristics, is often used for image deblurring~\cite{fang2023self,fergus2006removing}, it can also be used to infer the spatial motion kernels under strong assumptions~\cite{chen1996image}.
Later work improved performance by leveraging deep learning trained on annotated blur patterns, which are synthetically generated from per-pixel PSFs~\cite{gong2017motion}.
Similarly, subsequent methods exploit spatially non-uniform kernels to solve for complex motion blur kernels~\cite{kohler2012recording,fang2023self}.
However, it has been observed that the blur kernel is not able to model occlusions, unless additional temporal weights are added to model the foreground and background components~\cite{brooks2019learning}.
In other words, the existing methods heavily rely on PSF modeling, thereby restricting their applicability primarily to motion-blur removal tasks. 
The closely related work from~\cite{argaw2021optical} learns pixel motion from a single blurred image by exploiting latent two-view flow representations. 
Given one blurred image, \cite{argaw2021optical} predicts the first and last latent feature maps and subsequently regresses flow vectors similarly to the conventional two-frame optical flow estimation.
In this paper, we exploit pixel-wise motion estimation to recover the global camera motion during the exposure time, from a robust geometric perspective.

\vspace{-5pt}
\subsection{Estimation of Epipolar Geometry}
\label{subsec:related_geometry}

The fundamental matrix~\cite{luong1992matrice,luong1996fundamental} captures the geometric relation between two uncalibrated cameras observing a single rigid scene with unknown intrinsic parameters.
It defines the epipolar constraint between corresponding points across two images~\cite{huang1994motion,schonberger2016structure,crandall2011discrete}, which is crucial for the VO and SfM tasks. 
As a result, the fundamental matrix has been widely used to supervise pairwise correspondence learning~\cite{wang2020learning,jafarian2018monet}. 
The epipolar distances between points and their corresponding epipolar lines, computed from the fundamental matrix across the image, have been used as geometric constraints for learning human and animal keypoints~\cite{jafarian2018monet}.
The fundamental matrix is classically computed for two-view correspondences or an image pair~\cite{zhang1998determining,wexler2003learning}.
Recently, a learning-based approach~\cite{ranftl2018deep} was proposed to regress the matrix, supervised by weighted-putative correspondences from two images. 
The work in \cite{ranftl2018deep} found that directly regressing fundamental matrix entries without geometric cues leads to very poor performance.
The aforementioned approaches~\cite{zhang1998determining,wexler2003learning,ranftl2018deep} are only applicable to the classical two-view setting, and they cannot be directly applied to a single motion-blurred frame, where the geometric relation between the camera poses at shutter opening and closing must be inferred without access to two instantaneous images.
This single-frame setting introduces the additional challenge of the temporal direction ambiguity of each smear path.
Solving geometry from motion blur further enables practical applications such as motion segmentation, which separates locally blurred objects from the globally blurred scene.
\vspace{-5pt}
\section{Methodology}
\label{sec:method}
We present our proposed method in detail. 
\cref{subsec:smear-est} outlines the data preparation, smear annotations, and our representation to improve the robustness of smear learning.
\cref{subsec:epipolar-est} presents the epipolar geometry and explains how the fundamental matrix is robustly estimated from a single motion-blurred image.

\subsection{Smear Estimation}
\label{subsec:smear-est}

\subsubsection{Motion-Blurred Image Curation.} 
A motion-blurred image can be modeled as the integration of an instantaneous image $X(t)$ over the camera exposure interval $[t_o,t_c]$ plus sensor noise~\cite{chen1996image}. 
Hence, the blurry image can be approximated by averaging instantaneous samples.
Similar to~\cite{argaw2021optical,purohit2019bringing,chen2025imageimu,Zhong_2023_CVPR}, we obtain the blurry image $Y$ by averaging $N$ intermediate sharp images $X_k$ and adding noise $\mathcal{N}$:
\begin{equation}
    Y = \int^{t_c}_{t_o} (X(t)+\mathcal{N}(t))dt 
    \approx \frac{1}{N} \sum_{k=1}^{N} X_k + \mathcal{N}~.
    \label{eq:blur_approx}
\end{equation}
Following~\cite{argaw2021optical}, we use the synthetic Monkaa and Driving datasets~\cite{mayer2016large}, which provide pairs of sharp frames with ground truth optical flows $\mathbf{f} \in \mathbb{R}^{H\times W\times 2}$.
To obtain smooth motion-blurred patterns, we adopt the interpolation network~\cite{jiang2018super} with the input of two consecutive sharp images coupled with the provided optical flow to generate the intermediate frames $X_k$ used in~\cref{eq:blur_approx}. 
The number of generated frames needed for smoothness is determined from the maximum magnitude of the optical flow. 
Particularly, we set the number of frames to $N=\max(2s_\textup{max}, 15)$, where $s_\textup{max}=\max_{i,j}(|\mb{f}(i,j)|)$, to satisfy the sampling theorem \cite{sonkahlavacboyle2008}.
We apply the same data curation procedure for the hybrid-real motion-blur GoPro dataset~\cite{nah2017deep}.
We also use the RBI dataset~\cite{Zhong_2023_CVPR}, which provides real motion-blurred images, and sharp frames captured under high-frame-rate and fast camera motions.

\subsubsection{Smear Path Annotations.} 
A smear path can be approximated as a line~\cite{argaw2021optical,gong2017motion}, which is sufficiently expressive to model the blurry patterns~\cite{brooks2019learning}. 
The two endpoints of the line are determined from the ground truth optical flows. 
Hence, a motion-blurred smear path in an image plane is represented as a 2D vector positioned at the middle of the line.
However, smear paths are not always valid at all pixels due to potential occlusions during exposure time or pixels moving out of frame~\cite{argaw2021optical} (see Suppl.~\cref{subsec:supp_crosscheck}).
To distinguish the valid blurry patterns from the invalid ones, we use the cross-check procedure~\cite{baker2011database,menze2015object} to check the forward-backward flow consistency between the first and last frames, formulated as:
\begin{equation}
    \label{eq:cross}
    \begin{split}
        & d^{\textup{cr}}(i,j) = \lVert (i,j)^\top - \mb{C}_{\textup{bw}}(\mb{C}_{\textup{fw}}(i,j)) \rVert + \lVert \mb{C}_{\textup{fw}}(\mb{C}_{\textup{bw}}(i,j)) - (i,j)^\top \rVert~, \\
        & m^{\textup{cr}}(i,j) = 
            \begin{cases}
                1 & \text{if } d^{\textup{cr}}(i,j) \leq \varepsilon_{\textup{cr}}, \\
                0 & \text{otherwise.}   
            \end{cases}
    \end{split}
\end{equation}
where $d^{\textup{cr}}\in\mathbb{R}^{H \times W}$ is the cross-check distance map and $(i,j)^\top$ is the pixel coordinate at the position in the image plane; ${\mb{C}}(.)$ denotes the pixel coordinate warping process according to the ground truth forward ${\mb{f}}_{\textup{fw}}$ and backward ${\mb{f}}_{\textup{bw}}$ optical flows.
If the cross-check distance is smaller than a user-defined threshold $\varepsilon_{\textup{cr}}$, the blurry pattern is deemed a reliable motion cue. 
Using the map $m^{\textup{cr}}$ is novel compared to~\cite{argaw2021optical}, and as we demonstrate in the experiments shown in ~\cref{fig:ablation-fig}, it helps the network learn to interpret blurry patterns by ignoring the irregular ones.

\subsubsection{Smear Representation and Learning.}
\begin{figure}[t]
    \centering
    \includegraphics[width=0.98\linewidth]{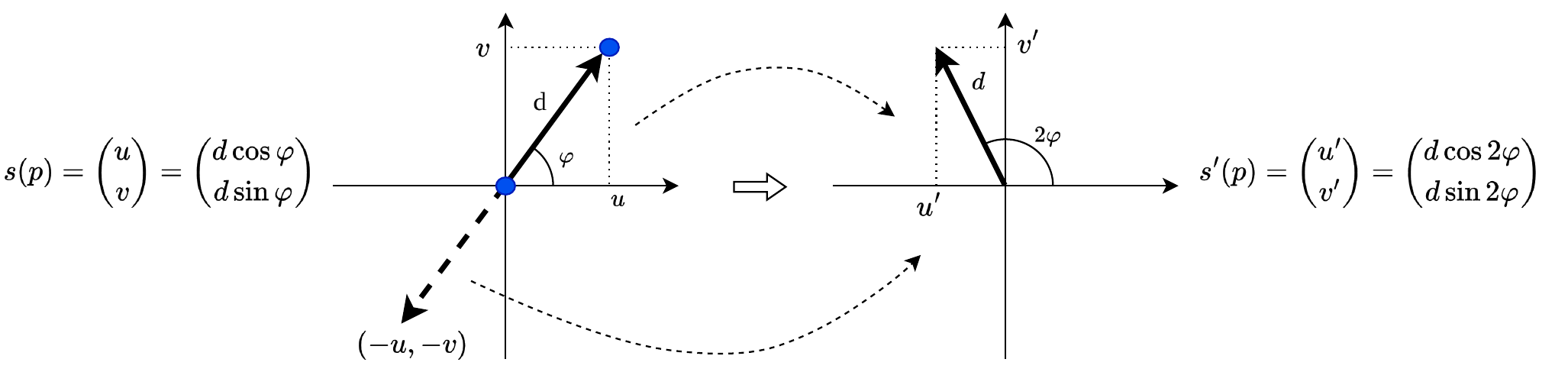}
    \caption{
    Double angle-based representation for smear paths in the motion-blurred image.
    Two local vectors $(u, v)$ and $(-u, -v)$ (left), express the same smear pattern, are transformed to a unique vector $s'(p)=(u', v')$ (right) to avoid the ambiguity.
    }
    \label{fig:double_vs_local}
\end{figure}
The flow~\cite{sun2018pwc} is described as local vectors ${\mb{f}}(i,j)=(u,v)^\top$ where $u$ and $v$ are the horizontal and vertical components, respectively. 
In the case of single-frame estimation, an ambiguity exists (\cref{fig:double_vs_local}), wherein a motion vector can be equivalently represented as either $(u,v)$ or $(-u,-v)$.
To address this issue, \cite{gong2017motion} constrains the $u$-component to be positive, whereas $u$ and $v$ are directly regressed regardless of value-sign in~\cite{argaw2021optical}. 
In this work, we use the double-angle representation \cite{bigun87} of the smear vector. 
This replaces the two equivalent cases $(u,v)$ and $(-u,-v)$ with a single vector $(u', v')$. 
Particularly, vectors $(u, v)$ with angle $\varphi$ and $(-u, -v)$ with $\varphi+\pi$ are both converted into a single vector $(u', v')$ that retains the same vector magnitude but with an orientation of $2\varphi$.
See~\cref{fig:double_vs_local} for a visual illustration of the conversion.

In the first stage of our method, we output smear predictions coupled with an uncertainty measurement. 
We argue that the additional uncertainty is a crucial complement for two reasons:
1) blurry patterns do not always exist at all pixels, requiring the network to have the ability to distinguish valid/invalid smear predictions; and 2) the second stage heavily depends on the predicted smears, making it preferable to optimize the fundamental matrix over reliable predictions rather than all smears.
A straightforward way to incorporate uncertainty measures is to use the Gaussian negative log-likelihood loss:
\begin{equation}
    \mathcal{L}_R(u',v',w) = \frac{(u'-u'^*)^2+(v'-v'^*)^2}{2 \sigma^2} + \log(\sigma^2)~,
    \label{eq:loss_no_cr}
\end{equation}
where $(u'^*, v'^*)$ is the ground-truth double-angle representation.
Inspired by~\cite{lakshminarayanan2017simple}, we pass the third model output, $w$, through the \emph{softplus} function to ensure a positive standard deviation $\sigma=\textup{log}(1+\textup{exp}(w))$.
In order to make use of the cross-check map in~\cref{eq:cross}, we extend the loss using the following expression:
\begin{equation}
\mathcal{L}(u',v',w) = m^{\textup{cr}} \mathcal{L}_R + \textup{ReLU}\left(\frac{1-m^{\textup{cr}}}{\sigma^2}-\alpha\right)~,
    \label{eq:loss}
\end{equation}
where $m^{\textup{cr}}$ is the ground-truth cross-check map.
The first term in~\cref{eq:loss} supervises the smear vector and its variance at valid pixels, while the second term aims to increase variance at invalid pixels.
The valid and invalid pixels are defined in $m^{\textup{cr}}$.
We use a $\textup{ReLU(.)}$ function along with a small margin $\alpha$ to avoid unbounded growth of the variance.
During inference, the outputs of double-angle-based smears $s'=(u',v')$ are converted back to local orientation $s=(u,v)$ as in \cref{eq:convert_to_local}.
The detailed mathematical derivation can be found in Suppl.~\cref{sec:supp_double_angle}.

\begin{equation}
    s=\begin{pmatrix}u\\v\end{pmatrix}= \sqrt{u'^2+v'^2} \begin{pmatrix}
        \cos (0.5 \tan^{-1}(v', u')) \\
        \sin (0.5 \tan^{-1}(v', u'))
    \end{pmatrix}.
    \label{eq:convert_to_local}
\end{equation}

%
\subsection{Geometry from Motion Blur}
\label{subsec:epipolar-est}

\subsubsection{Preliminaries.}
In regular epipolar setups~\cite{hartley_zisserman2004}, two cameras look at the scene with camera projection matrices $C_i=K_i[R_i| \;t_i]; \; i\in\{1,2\}$, where $\{K_i, R_i, t_i\}$ are the intrinsic parameters, rotation matrix and translation vector, respectively. 
The fundamental matrix~\cite{luong1996fundamental} encapsulating the relative geometric relationship between two cameras, is defined as ${\mb{F}} = [{\mb{e}}_2]_\times C_2C_1^\dag$, where ${\mb{e}}_2$ denotes the first camera epipole. 
Any valid correspondence $x \leftrightarrow y$, should satisfy the epipolar constraint:
\begin{equation}
    \label{eq:orig-epip-const}
    x^\top {\mb{F}}y = 0~.
\end{equation}
Such a 7-DoF fundamental matrix is classically obtained by the 8-point (8P) or the 7-point algorithm~\cite{hartley_zisserman2004}, or a maximum likelihood estimator, \ie the gold standard (GS) method~\cite{hartley_zisserman2004}.
These methods cannot be used in single-frame setups due to unknown temporal direction and we address this challenge via GeoBlur.
%
\subsubsection{Solving Motion-blurred Geometry.}
Each smear path in a blurry image is the integration of projection of a single 3D point.
This means two endpoints of a smear should satisfy the epipolar constraint in~\cref{eq:orig-epip-const} if they follow the global scene motion. 
In the single-frame context, the correspondence could be extracted from the start and end points of a smear line acquired from the first stage of our method. 
A problem emerges at this point; it is so far not possible to predict the motion sign from a single blurred image. 
A correct correspondence can thus satisfy either of the following two cases:
\begin{gather}
    (\bar{p} - \bar{s}(p))^\top \mathbf{F} (\bar{p} + \bar{s}(p)) = 0~, \textup{ or}
    \label{eq:F_constraint_a} \\
    (\bar{p} + \bar{s}(p))^\top \mathbf{F} (\bar{p} - \bar{s}(p)) = 0
    \; \; \Leftrightarrow \; \; (\bar{p} - \bar{s}(p))^\top \mathbf{F}^\top (\bar{p} + \bar{s}(p)) = 0~,
    \label{eq:F_constraint_c}
\end{gather}
where $\bar{p} = (p^\top \: 1)^\top$ is the homogeneous coordinates of $p$ and $s(p)=(u,v)$ is the predicted smear vector at $p$. 
Similarly, $\bar{s}(p) = (s(p)^\top \: 0)^\top$ is the homogeneous version of $s(p)$. 
The ambiguities in~\cref{eq:F_constraint_a}~and~\cref{eq:F_constraint_c} mean that conventional methods, \eg the 8-point algorithm and the gold standard method~\cite{longuet1981computer}, are no longer applicable.
Each of the two endpoints of a smear path should satisfy either~\cref{eq:F_constraint_a}~or~\cref{eq:F_constraint_c}.
Therefore, we treat the fundamental matrix estimation as the following constrained optimization problem over 7 smear paths:
\begin{equation}
    \label{eq:fund_matrix_est}
    \begin{split}
        & {\mb{F}} = \arg\min_\mb{F}(\lVert \min(|\mb{M} \textup{vec}(\mb{F})|, |\mb{M} \textup{vec}(\mb{F}^\top)|) \rVert^2)~, \\
        & \textup{s.t.}~\det(\mb{F})=0, \lVert \mathbf{F} \rVert=1~\textup{and}~
        \mathbf{M}=
        \begin{pmatrix}
        	(\bar{p}_1 - \bar{s}({p_1})) \otimes (\bar{p}_1 + \bar{s}({p_1})) \\
        	\vdots	\\
        	(\bar{p}_7 - \bar{s}({p_7})) \otimes (\bar{p}_7 + \bar{s}({p_7})) 
        \end{pmatrix},
    \end{split}
\end{equation}
where $\min(.,.)$ performs element-wise min, $\otimes$ denotes \emph{Kronecker product}, and $\bar{s}(p)$ is the smear vector at pixel $p$ in homogeneous coordinates.
In the matrix $\mb{M}\in\mathbb{R}^{7\times9}$, the $\bar{p}_1, \ldots, \bar{p}_7$ denote the homogeneous coordinates of points in the 7-point subset.
Due to the ambiguity of the global motion direction in a single image, the true $\mb{F}$ and a correspondence satisfy either~\cref{eq:F_constraint_a}~or~\cref{eq:F_constraint_c}.
Hence, minimizing~\cref{eq:fund_matrix_est} can result in either $\mb{F}$ or $\mb{F}^\top$.

\begin{figure}[t]
\centering
    \begin{subfigure}{.3\linewidth}
        \includegraphics[width=\textwidth]{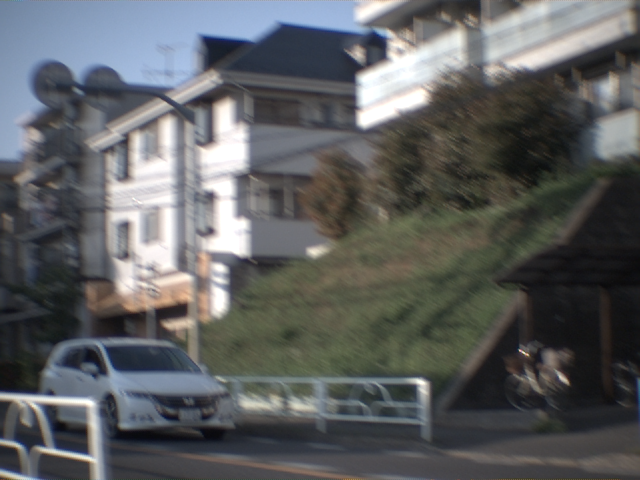}
    \end{subfigure}
    \begin{subfigure}{.3\linewidth}
        \includegraphics[width=\textwidth]{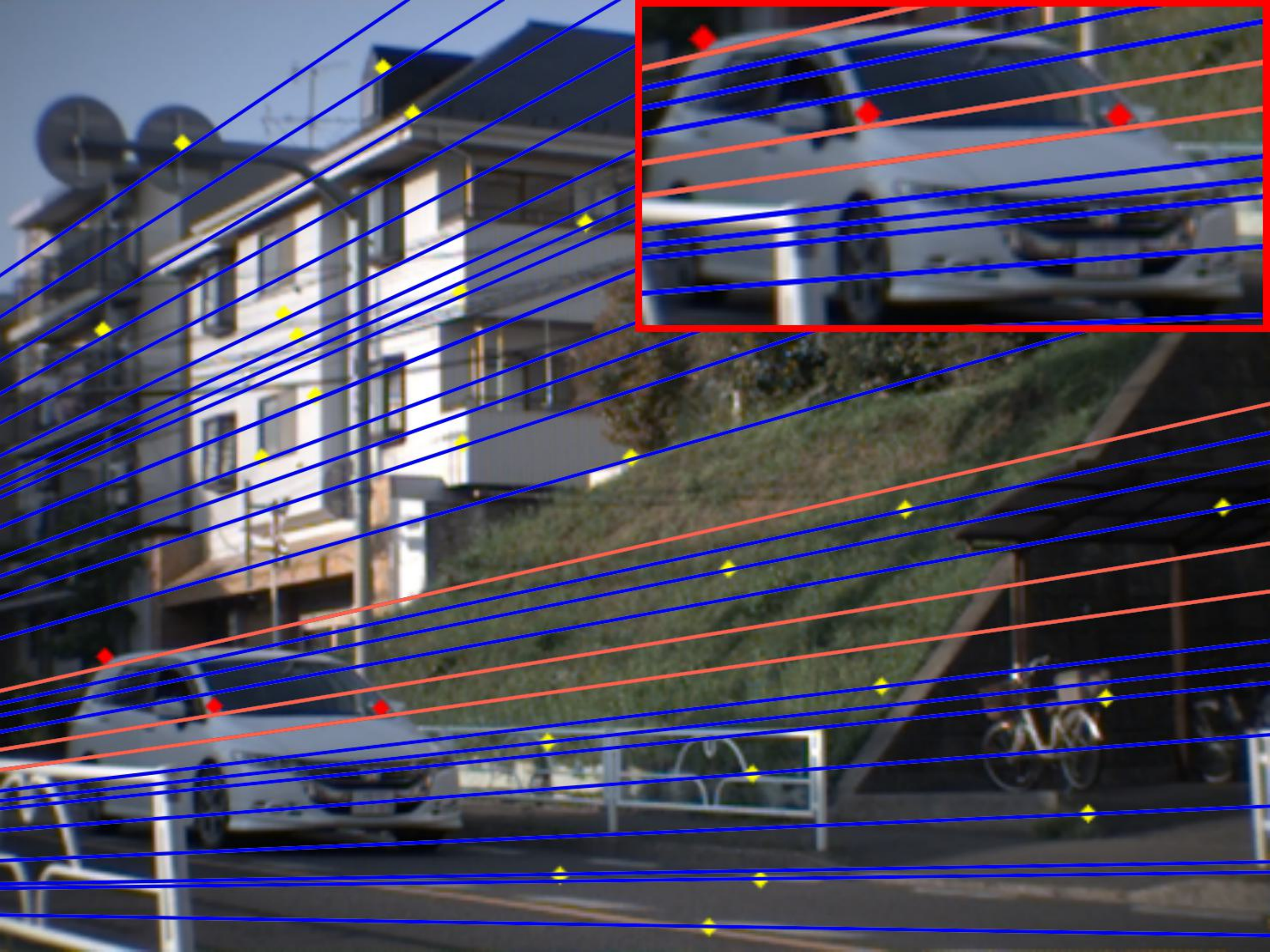}
    \end{subfigure}
    \begin{subfigure}{.3\linewidth}
        \includegraphics[width=\textwidth]{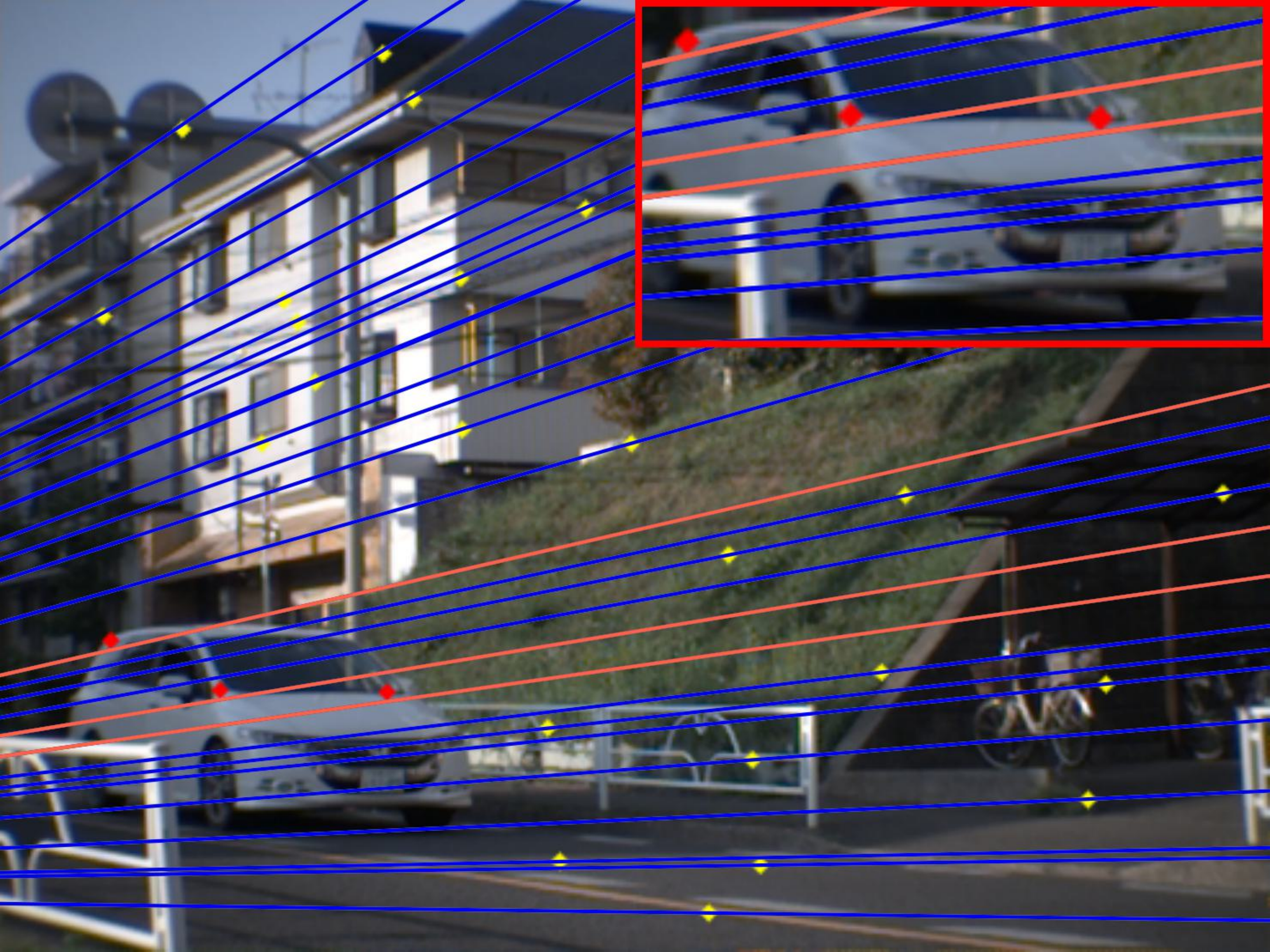}
    \end{subfigure}
    \caption{
    An example of GeoBlur's robustness to local motion.
    Left: blurry input; middle and right: epipolar lines computed from the estimated fundamental matrix, projected onto the sharp images at the start and end of the exposure interval.
    The epipolar lines (\textcolor{blue}{blue}) align with global correspondences (\textcolor{goldenrod}{yellow}), indicating good global geometry estimation.
    Meanwhile, the local points (\textcolor{red}{red} dots on the moving car) do not lie on the epipolar lines (\textcolor{red}{red}) demonstrate our method’s robustness to local object motion.
    }
    \label{fig:robust}
\end{figure}
\vspace{-0.3em}
\subsubsection{Robust Estimation.}
The first stage provides per-pixel predictions of smear vectors, $s=(u,v)^\top$, and their uncertainties $\sigma$.
This allows us to select a fraction $\mathcal{S}_\beta$ of the predicted smear vectors, using a threshold on $\sigma$. 
This gives a set of pixel coordinates $\mathcal{P}_\beta = \{p_1,...,p_L\}$ along with smear vectors $\mathcal{S}_\beta = \{s_1,...,s_L\}$, for which $\beta=L/(W\times H)$, and $W, H$ are the image size.
Besides suppressing inaccurate smear predictions, the threshold provides a means of smear selection to facilitate optimization.
This selection has a sweet spot, as a larger $\beta$ can increase the likelihood of containing outliers, while a small subset may fail to adequately represent the global motion, see~\cref{subsec:ablation} for further investigation.
For robustness to other motion components in the scene, we use the \emph{Preemptive RANSAC}~\cite{nister2005preemptive}, with~\cref{eq:fund_matrix_est} as the minimal solver on 7-point subsets, drawn from $\mathcal{P}_\beta$, and define the consensus set by thresholding the \emph{Sampson Error}~\cite{luong1996fundamental,sampson1982fitting} on the rest of $\mathcal{P}_\beta$ with a threshold $\tau_{SE}$.
Additionally, our smear vectors have directional ambiguity while the \emph{Sampson Error} requires a correspondence.
Hence, we use a symmetrized version of the Sampson error, denoted as \SErrMin, and expressed as:
\begin{gather}
    \textup{SErr-S}(p) = \min( \varepsilon^2_\textup{samp}(\bar{p},\bar{s}(p),\mb{F}), \varepsilon^2_\textup{samp}(\bar{p},\bar{s}(p),\mb{F}^\top))~,~\textup{and}
    \label{eq:sampson_error}\\
    \varepsilon^2_\textup{samp}(\bar{p},\bar{s}(p), \mb{F}) = \frac{|(\bar{p} - \bar{s}(p))^\top \mb{F} (\bar{p} + \bar{s}(p))|^2}{l_1^2+l_2^2+{l'_1}^2+{l'_2}^2}~,
    \label{eq:sampson}
\end{gather}
where $l, l'$ are epipolar lines computed from the estimated ${\mb{F}}$~\cite{hartley_zisserman2004}.
This makes our approach robust to scenes with local motion, where smears are produced by independently moving objects during exposure; an example is demonstrated in~\cref{fig:robust}.
The \SErrMin reflects the true \emph{Sampson Error} as \SErrMin is zero only if the true $\mb{F}$ or $\mb{F}^\top$ is given.
The start-end assignment is selected as the smaller value of the two error elements in~\cref{eq:sampson_error}.


\section{Experiments}
\label{sec:experiments}

\subsection{Datasets}
\label{subsec:datasets}
We evaluate on four datasets: a real-world motion blur dataset (RBI)~\cite{Zhong_2023_CVPR}; a hybrid-realistic dataset (Gopro)~\cite{nah2017deep}; and two synthetic datasets from~\cite{mayer2016large} (Monkaa and Driving).
RBI is a real-world dataset providing real motion-blurred images and corresponding sharp images.  
It uses a beam-splitter setup, where one camera captures a long-exposure blurred image and the other simultaneously records high-frame-rate sharp images over the same field-of-view.
However, this dataset contains only a few multiple-motion scenes, so we further evaluate our method on three more challenging dynamic datasets.
GoPro is a hybrid realistic dataset, where motion blur is synthesized by averaging high-FPS sharp frames from real scenes.
For GoPro and RBI, we use~\cite{sun2018pwc} on the first and last sharp frames to obtain pseudo ground-truth smear paths, and then apply the GS method~\cite{hartley_zisserman2004} to estimate pseudo ground-truth fundamental matrices.
Driving and Monkaa are synthetic datasets with ground-truth camera intrinsics and extrinsics, from which we compute ground-truth fundamental matrices. 
Using the provided ground-truth optical flow, we generate intermediate frames using~\cite{jiang2018super} and average them to synthesize motion blur.

\subsection{Evaluation metrics}
\label{subsec:metrics}
\noindent \textbf{Reprojection metrics}.
Since directly comparing the matrix entries with the ground truth is not meaningful~\cite{zhang1998determining}, we instead use geometric consistency measurement. 
A common practice is to measure the percentage of ground-truth correspondences that are consistent with the estimated fundamental matrix~\cite{ranftl2018deep,zhang1998determining,hartley_zisserman2004}.
To determine the extent to which the estimation fits the global motion, we calculate the percentage of ground-truth correspondences (smear paths) that fall below a threshold on the \SErrMin, measured in~\cref{eq:sampson_error}.
We report this percentage as the inlier ratio, together with the median distance \SErrMin.

\noindent \textbf{Relative pose metrics}.
Given a known intrinsic matrix, the relative camera pose can be recovered from the fundamental matrix and compared with the ground truth. 
Following~\cite{edstedt2024roma,wang2024dust3r}, we report Relative Rotation Accuracy (RRA)and Relative Translation Accuracy (RTA), measured in degrees. We also report the AUC computed from both rotation and translation errors. Following \cite{edstedt2024roma}, we report the AUC at the threshold of $20^{\circ}$ (AUC@20). 

\noindent \textbf{Smear metric}.
The metric is defined by end-point-error (EPE) between the estimated smear vector $s(p)=(u,v)$ and the ground-truth $s^{gt}(p)=(u^{gt}, v^{gt})$.
Due to the per-pixel ambiguity of the motion direction of a single input, for a fair evaluation, we replace the plain EPE with a sign-agnostic version:
\begin{equation}
\textup{EPE-S}=\textup{min}\{\textup{EPE}(s(p), s^{gt}(p)), \; \textup{EPE}(-s(p), s^{gt}(p))\}\,.
\label{eq:epe-s}
\end{equation}

\subsection{Baselines}
\label{subsec:baselines}
We compare our approach with both direct and indirect methods.
The direct methods estimate the fundamental matrix or relative pose directly from a single blurred input, without using temporal priors.
First, we use the Image-as-IMU~\cite{chen2025imageimu}, which can estimate relative pose from a single input.
We recover the fundamental matrix using the known intrinsics, and fine-tune the model using the same dataset settings as ours.
Second, we design a straightforward regression-based baseline that directly regresses the fundamental matrix from a single blurred input, inspired by~\cite{ranftl2018deep}.
We use a shifted-window Transformer~\cite{Liang_2021_ICCV}, combined with a residual multi-scale feature design as the backbone.
In contrast to direct methods, indirect methods estimate the temporally ordered instantaneous frames from the blurred input using a known temporal prior, then apply conventional epipolar geometry to the first and last frames.
We use the method of~\cite{Zhong_2023_CVPR}, namely Interp+8P, which takes three blurred images as input—the previous, current, and next frames—and interpolates nine intermediate sharp frames. 
The interpolated frames are temporally ordered. We then apply the 8-point algorithm~\cite{hartley_zisserman2004} in a two-view setup using the first and last frames.

\subsection{Geometry Estimation}
\label{Geo Estimation}

\subsubsection{Fundamental matrix evaluation.}
\begin{figure}[t!]
\centering
\setlength{\tabcolsep}{0pt} 
    \begin{subfigure}{.19\linewidth}
        \includegraphics[width=\textwidth]{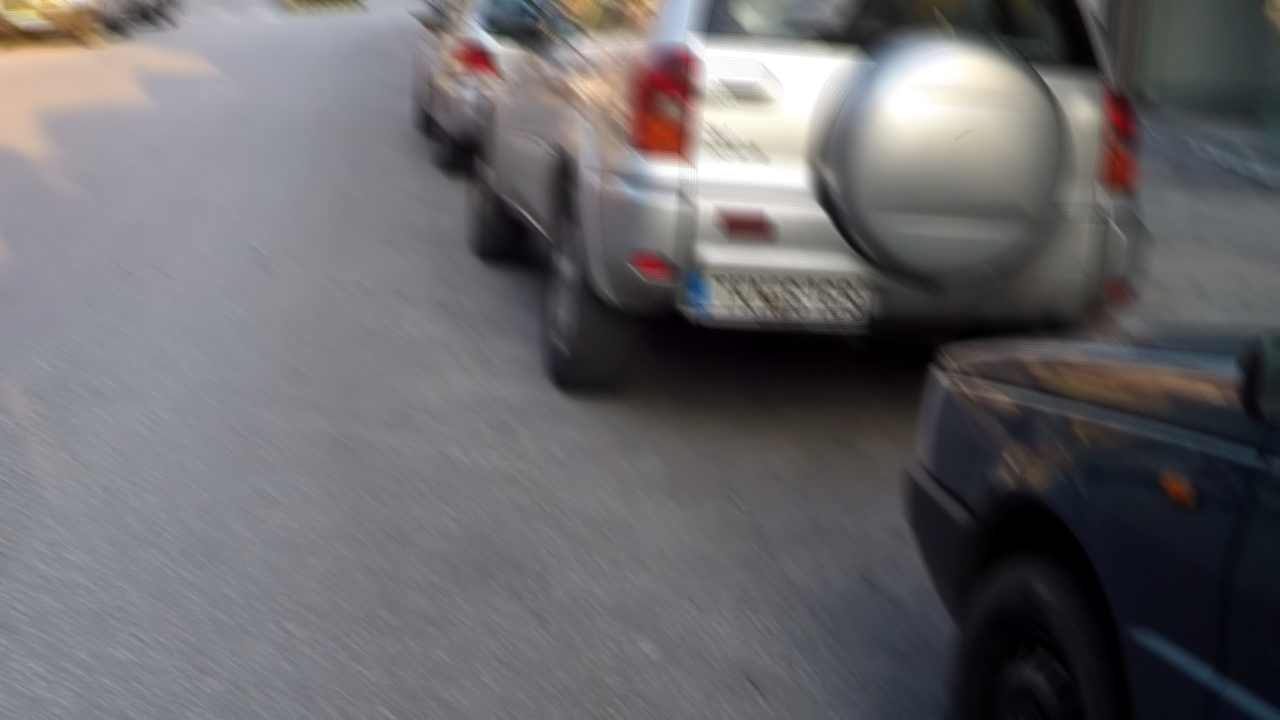}    
    \end{subfigure}\hspace{1pt}%
    \begin{subfigure}{.19\linewidth}
        \includegraphics[width=\textwidth]{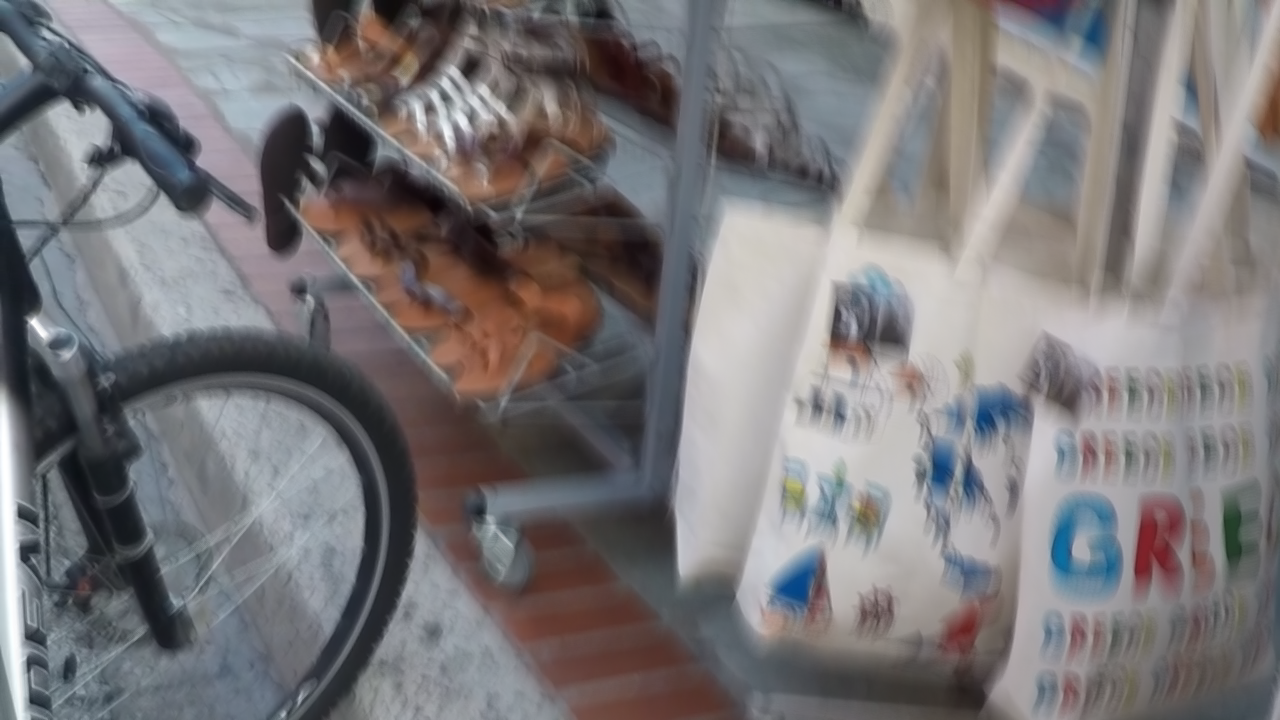}
    \end{subfigure}\hspace{1pt}%
    \begin{subfigure}{.19\linewidth} 
        \includegraphics[width=\textwidth]{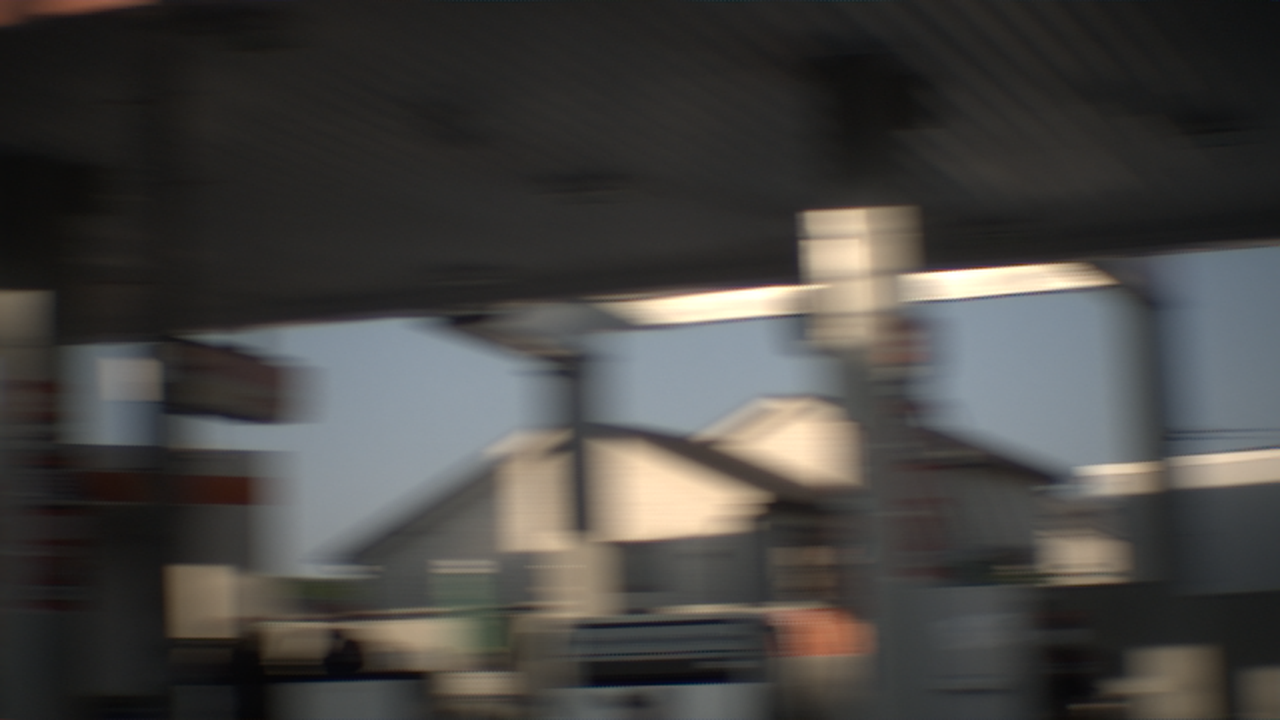}
    \end{subfigure}\hspace{1pt}%
    \begin{subfigure}{.19\linewidth} 
        \includegraphics[width=\textwidth]{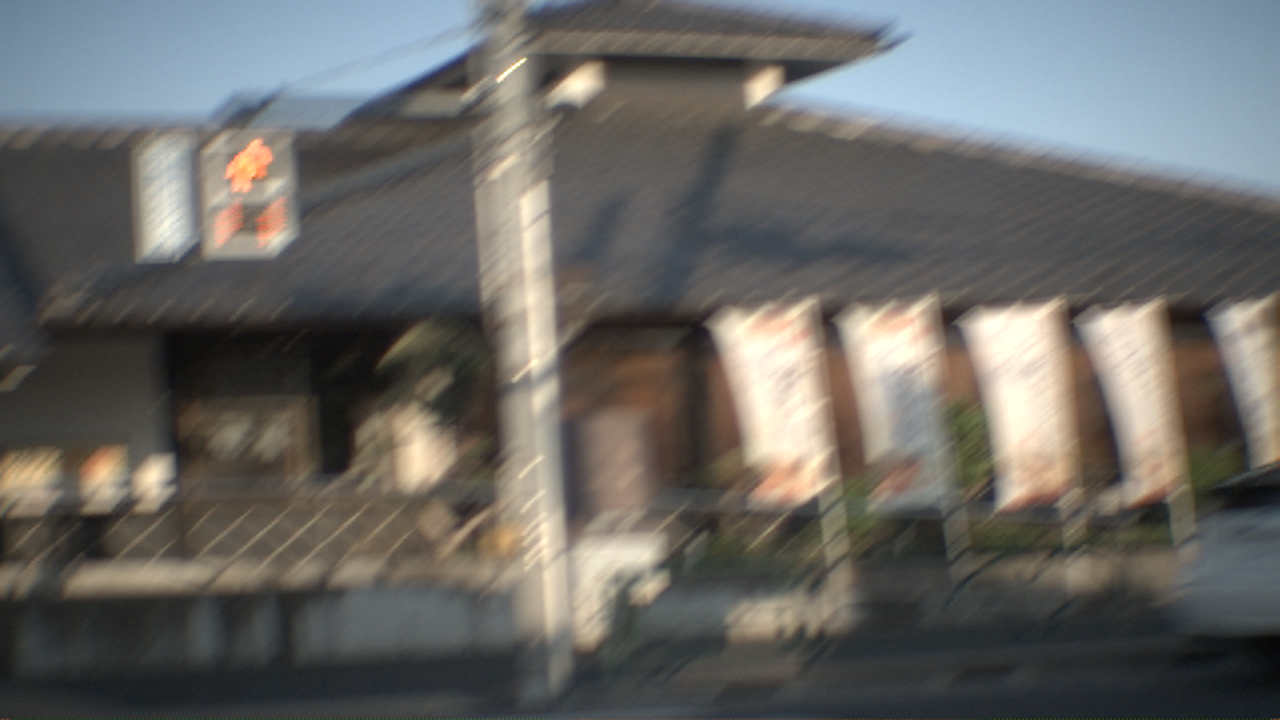}
    \end{subfigure}\hspace{1pt}%
    \begin{subfigure}{.19\linewidth}
        \includegraphics[width=\textwidth]{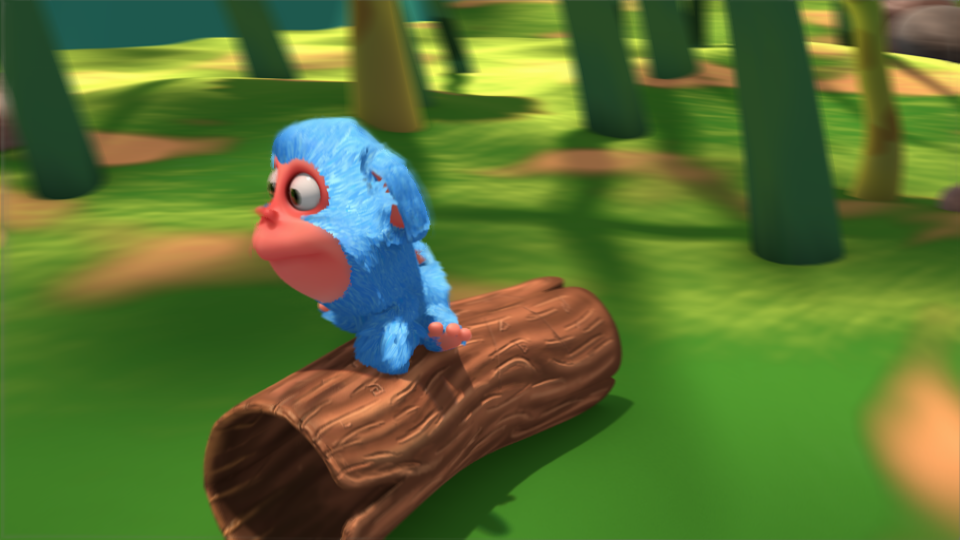}
    \end{subfigure}\hspace{1pt}%

    \begin{subfigure}{.19\linewidth}
        \includegraphics[width=\textwidth]{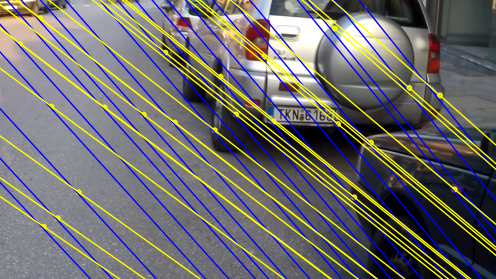}
    \end{subfigure}\hspace{1pt}%
    \begin{subfigure}{.19\linewidth}
        \includegraphics[width=\textwidth]{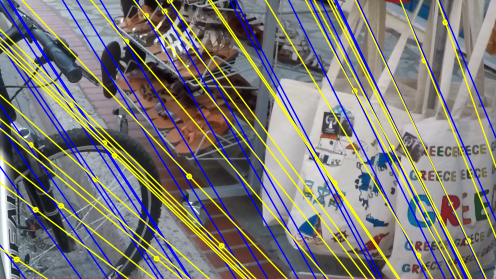}
    \end{subfigure}\hspace{1pt}%
    \begin{subfigure}{.19\linewidth}
        \includegraphics[width=\textwidth]{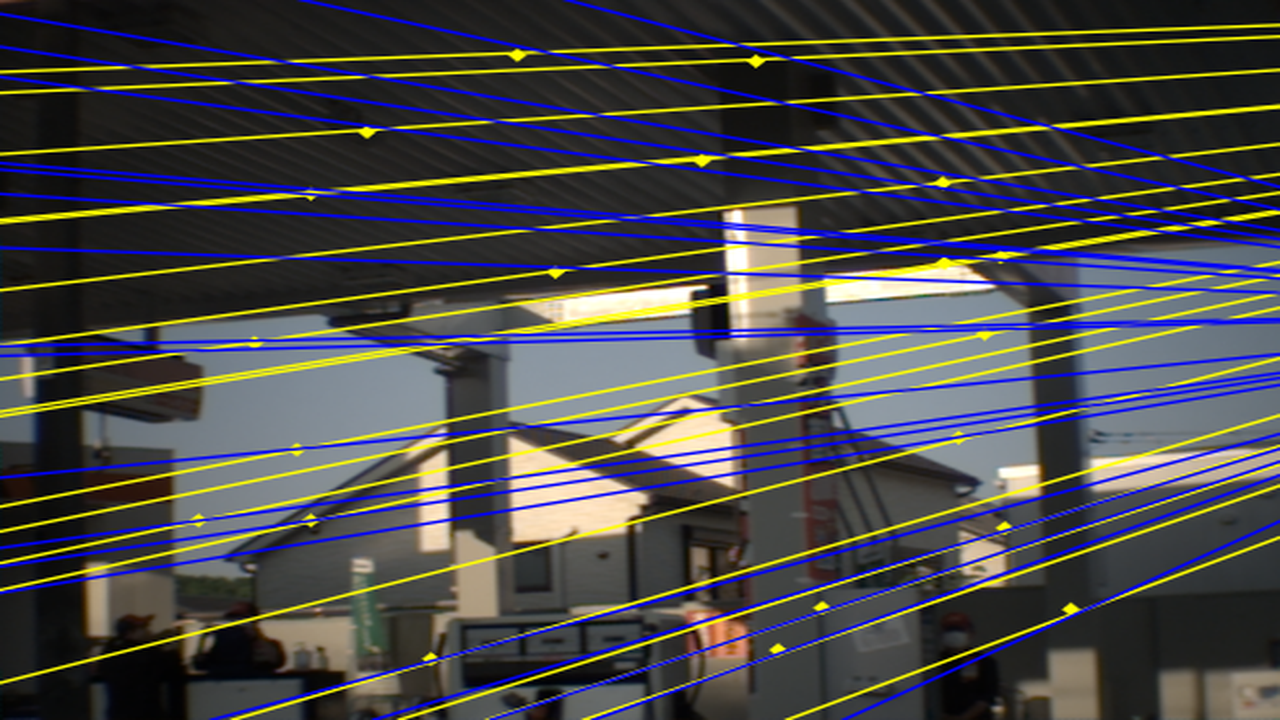}
    \end{subfigure}\hspace{1pt}%
    \begin{subfigure}{.19\linewidth}
        \includegraphics[width=\textwidth]{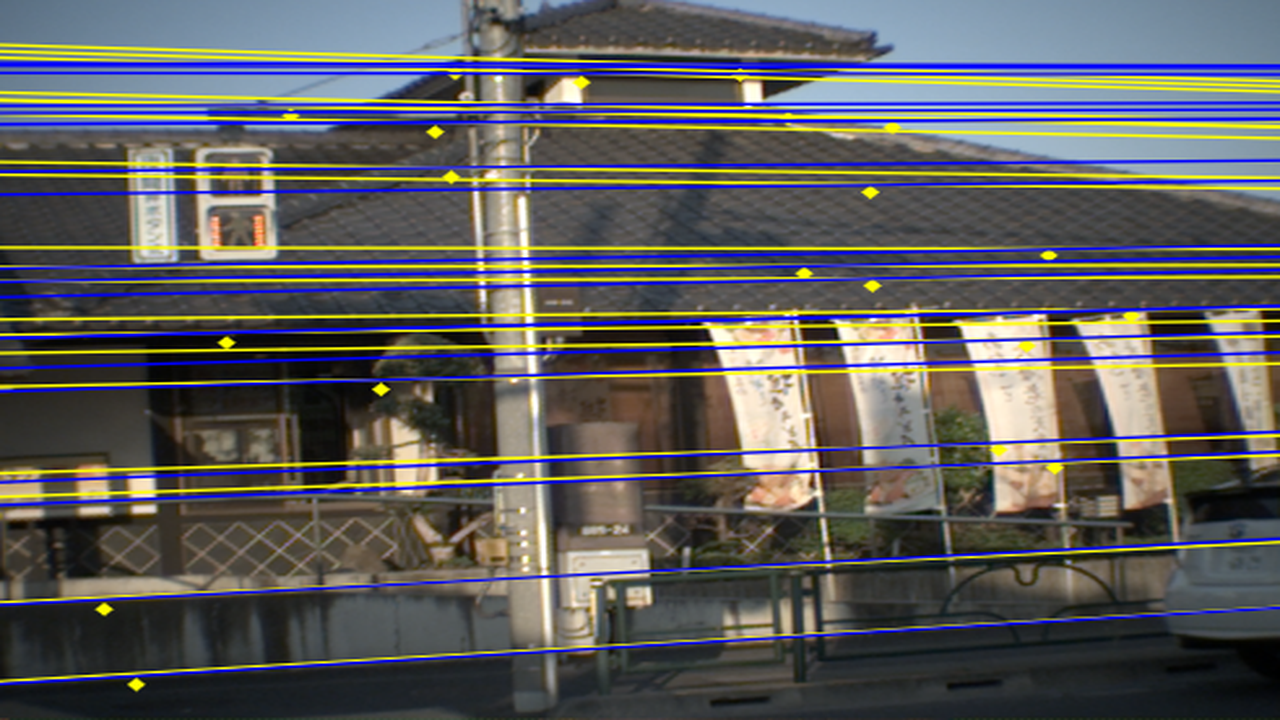}
    \end{subfigure}\hspace{1pt}%
    \begin{subfigure}{.19\linewidth}
        \includegraphics[width=\textwidth]{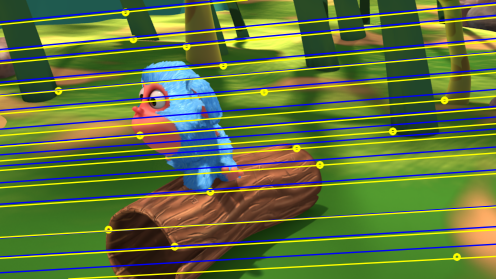}
    \end{subfigure}\hspace{1pt}%

    \begin{subfigure}{.19\linewidth}
        \includegraphics[width=\textwidth]{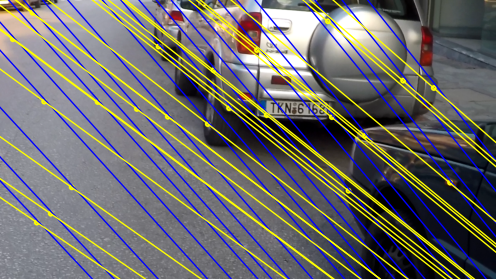}
    \end{subfigure}\hspace{1pt}%
    \begin{subfigure}{.19\linewidth}
        \includegraphics[width=\textwidth]{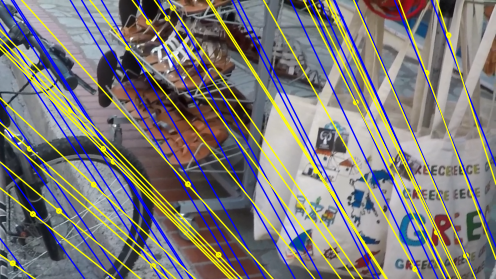}
    \end{subfigure}\hspace{1pt}%
    \begin{subfigure}{.19\linewidth}
        \includegraphics[width=\textwidth]{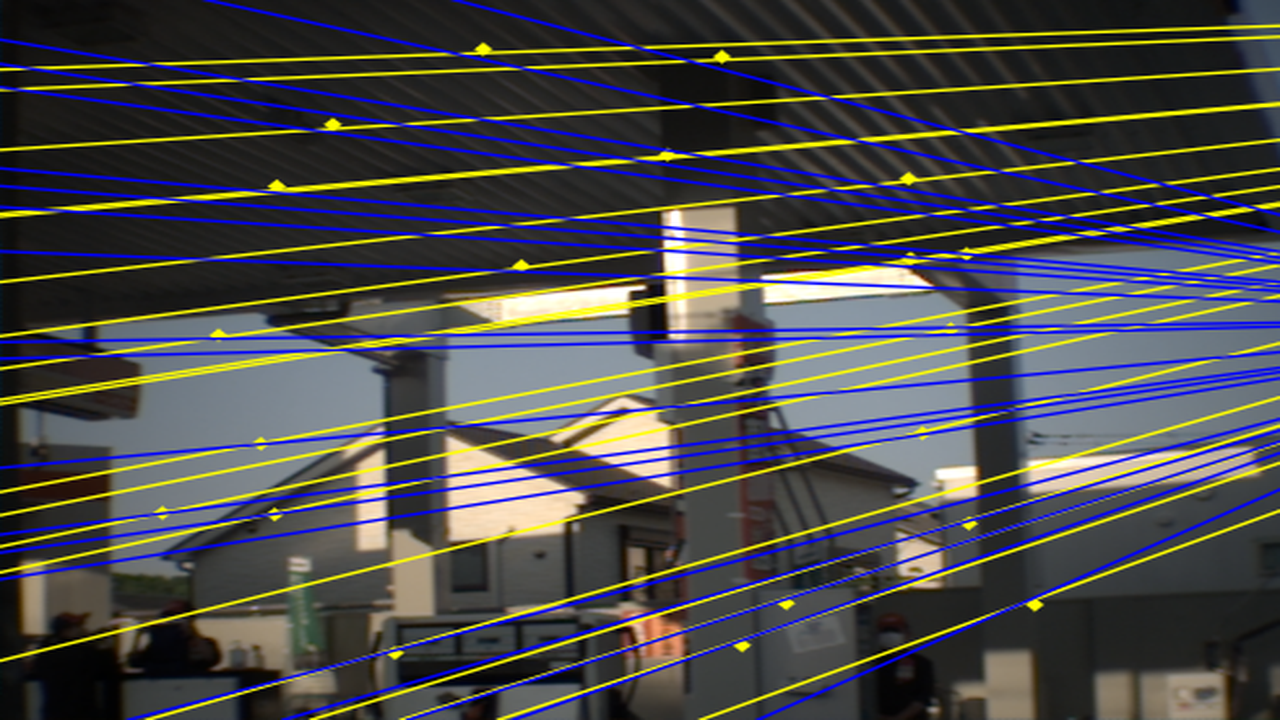}
    \end{subfigure}\hspace{1pt}%
    \begin{subfigure}{.19\linewidth}
        \includegraphics[width=\textwidth]{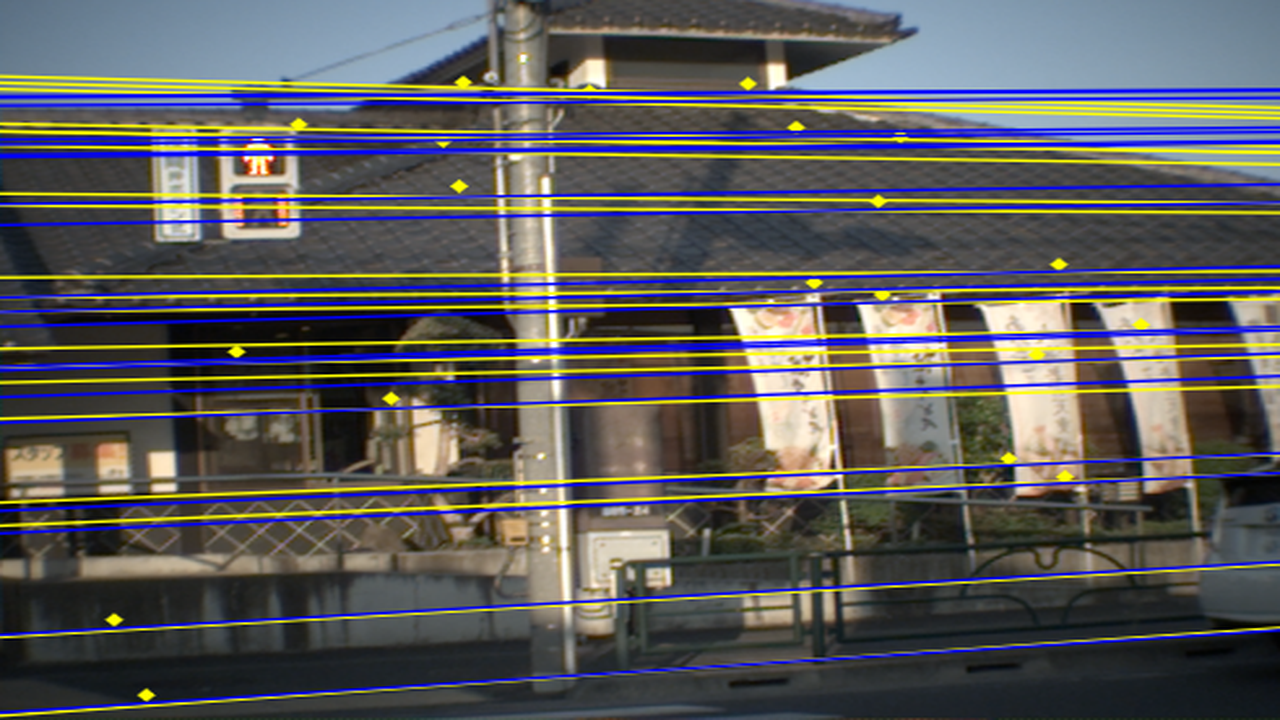}
    \end{subfigure}\hspace{1pt}%
    \begin{subfigure}{.19\linewidth}
        \includegraphics[width=\textwidth]{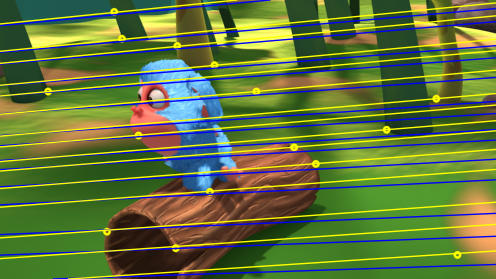}
    \end{subfigure}\hspace{1pt}%
    
    \caption{Qualitative examples of fundamental matrix estimations. Columns: 1$^{st}$-2$^{nd}$ (Gopro); 3$^{rd}$-4$^{th}$ (RBI); 5$^{th}$ (Monkaa). Row 1$^{st}$: blurry inputs. Row 2$^{nd}$ and 3$^{rd}$: the first and last (sharp) frames (only used for visualization). \textcolor{goldenrod}{Yellow} points are ground-truth correspondences.  \textcolor{goldenrod}{Yellow} and \textcolor{blue}{blue} epipolar lines are generated from ground-truth fundamental matrices and the estimated ones, respectively. (See Suppl. for more examples)
    }
    \label{fig:qualitative_results}
\end{figure}

We report the evaluation of the fundamental matrix estimation in~\cref{tab:fmatrix_inlier}.
Since the datasets contain dynamic scenes, where local object motions occur simultaneously with the global camera motion, not all ground-truth smear paths are consistent with the ground-truth fundamental matrix (\textbf{F}-Ground truth). 
Our method achieves significant improvements in the inlier-ratio metric over prior approaches on three datasets, with the only exception being the RBI dataset, where we reach the second best (39.74\%), on par with the best approach \textup{Image-as-IMU}~\cite{chen2025imageimu} (40.42\%).
This demonstrates the overall competitiveness and robustness across diverse motion-blur scenarios.
In addition,~\cref{fig:qualitative_results} visualizes the geometric properties of our method, providing qualitative evidence of its ability to recover epipolar geometry under motion blur.
It shows the epipolar lines, which correspond to the estimated geometry, on the virtual sharp frame planes.
These sharp frames are not available during inference and are used only for visualization purposes.
The epipolar lines align well with ground-truth correspondences, indicating that the geometric transformation is correctly estimated.
For example, the examples in $2^{nd}$~and~$3^{rd}$ columns in \cref{fig:qualitative_results}, the orientations of epipolar lines, projected from the estimated fundamental matrix, visually align with the ground truth and converge towards reasonably close points, reflecting accurate rotation and translation estimations. 

\begin{table}[t]
    \scriptsize 
    \caption{The evaluations of the fundamental matrix estimation. Here, inls is the percentage of inliers ($\%$), and med is the median of \SErrMin defined in \cref{eq:sampson_error}.
    }
    \label{tab:fmatrix_inlier}
    \centering
    \begin{tabularx}{\linewidth}{@{\extracolsep{\fill}}lcccccccccc}
        \toprule
        Method & direct & \multicolumn{2}{c}{Monkaa} & \multicolumn{2}{c}{Driving} & \multicolumn{2}{c}{GoPro} & \multicolumn{2}{c}{RBI} \\
        & & inls$\uparrow$ & med$\downarrow$ & inls$\uparrow$ & med$\downarrow$ & inls$\uparrow$ & med$\downarrow$ & inls$\uparrow$ & med$\downarrow$ \\ 
        \midrule
        Interp~\cite{Zhong_2023_CVPR}+8P~\cite{hartley_zisserman2004}  & $\times$ & 31.47 & 4.69 & 28.14 & 6.35 & 19.42 & 8.54 & 30.54 & 3.43 \\
        Image-as-IMU~\cite{chen2025imageimu} & {\checkmark} & 35.41 & 5.21 & 32.24 & 5.41 & 22.45 & 7.39 & \textbf{40.42} & 2.65 \\
        Regress-based & {\checkmark} & 21.64 & 3.78 & 22.24 & 7.36 & 21.14 & 5.58 & 18.45 & 7.29 \\
        GeoBlur (ours) & {\checkmark} & {\bf 45.02} & 2.87 & {\bf 32.79} & 4.68 & \bf{24.84} & 6.77 & 39.74 & 3.03 \\
        \midrule
        $\mb{F}$-Ground truth & & 60.45 & - & 70.28 & - & 58.10 & - & 59.39 & -\\
        \bottomrule
    \end{tabularx}
\end{table}

\begin{table}[t]
    \scriptsize
    \caption{The evaluations on relative pose metrics. AUC is reported at AUC$@20$.}
    \label{tab:rel_pose}
    \centering
    \begin{tabularx}{\linewidth}{@{\extracolsep{\fill}}lcccccccccccc}
        \toprule
        \multirow{2}{*}{Method} & \multicolumn{3}{c}{Monkaa} & \multicolumn{3}{c}{Driving} & \multicolumn{3}{c}{GoPro} & \multicolumn{3}{c}{RBI} \\
        & \tiny{RRA$\downarrow$} & \tiny{RTA$\downarrow$} & \tiny{AUC$\uparrow$} & \tiny{RRA$\downarrow$} & \tiny{RTA$\downarrow$} & \tiny{AUC$\uparrow$} & \tiny{RRA$\downarrow$} & \tiny{RTA$\downarrow$} & \tiny{AUC$\uparrow$} & \tiny{RRA$\downarrow$} & \tiny{RTA$\downarrow$} & \tiny{AUC$\uparrow$} \\ 
        \midrule
        Interp\cite{Zhong_2023_CVPR}+8P\cite{hartley_zisserman2004}  & 1.52 & 15.89 & 0.34 & 2.23 & 15.41 & 0.29 & 12.69 & 14.24 & 0.21 & 6.89 & 8.25 & 0.35 \\
        Image-as-IMU \cite{chen2025imageimu} & 6.56 & 7.58 & 0.45 & 5.54 & 11.24 & 0.27 & 4.45 & 9.87 & 0.29 & 3.25 & 5.07 & {\bf 0.41} \\
        Regress-based & 9.34 & 11.36 & 0.22 & 6.45 & 21.32 & 0.19 & 6.67 & 5.29 & 0.32 & 9.25 & 10.21 & 0.24 \\
        GeoBlur (ours) & 2.21 & 6.98 & {\bf 0.49} & 1.03 & 15.34 & {\bf 0.31} & 1.34 & 7.64 & {\bf 0.38} & 2.12 & 6.33 & {\bf 0.41} \\
        \bottomrule
    \end{tabularx}
    \vspace{-10pt}
\end{table}
\vspace{-0.5em}
\subsubsection{Relative Pose Evaluation.}
From the estimated fundamental matrix, the relative pose between the camera at shutter opening and shutter closing can be inferred given the intrinsic matrix. 
This decomposition yields four possible solutions, and the valid one is selected by triangulating the predicted smear-path endpoints~\cite{hartley_zisserman2004}. 
The correct solution is the one for which most triangulated points lie in front of both cameras.
We evaluate our method on various datasets, as shown in~\cref{tab:rel_pose}.
Our method outperforms prior direct and indirect methods in terms of AUC, which summarizes RRA and RTA of the estimated camera poses.
On the RBI dataset, our method achieves a competitive AUC$@$20 score of 0.41, matching the performance of \textup{Image-as-IMU}~\cite{chen2025imageimu} baseline, the most recent state-of-the-art method.
\cref{tab:fmatrix_inlier} and~\cref{tab:rel_pose} suggest that directly regressing the underlying camera motion from blurry patterns remains highly challenging.
The lower performance of \textup{Interp+8P} suggests that reformulating motion-blurred geometry as conventional two-view estimation is less effective.
In contrast, our method demonstrates that individual smear paths contain valuable geometric cues, enabling the recovery of the underlying global camera motion over the exposure time.
Note that all experiments use $\beta=45\%$, the best empirical value, unless otherwise stated.
Ablation study about the choice of $\beta$ is provided in~\cref{subsec:ablation}.

\begin{table}[t]
    \scriptsize     
    \setlength{\tabcolsep}{1.5pt}
    \caption{
    Smear path estimation EPE-S results using different smear representations.
    The EPE-S metric is defined in~\cref{eq:epe-s} and lower is better.
    }
    \label{tab:EPE-S_exp}
    \centering
    \begin{tabularx}{\linewidth}{@{\extracolsep{\fill}}lcccccc}
        \toprule
        Smear representation & Cross check & Monkaa & Driving & GoPro & RBI & Average \\ 
        \midrule
        local vectors~\cite{argaw2021optical}    & $\times$ & 6.13 & 27.66 & 7.94 & 11.33 & 13.26 \\
        restricted vectors~\cite{gong2017motion} & $\times$ & 3.57 & 21.23 & 7.22 & 10.18 & 10.55 \\
        double-angle vectors                     & $\times$ & 3.28 & 15.75 & 6.98 & 8.50  & 8.63  \\
        \midrule
        local vectors~\cite{argaw2021optical}    & \checkmark & 1.68 & 15.51 & 4.99 & 5.43 & 6.90\\ 
        restricted vectors~\cite{gong2017motion} & \checkmark & 1.18 & 13.31 & 3.15 & 4.12 & 5.44\\
        double-angle vectors                     & \checkmark & {\bf 0.72} & {\bf 9.12} & {\bf 0.95} & {\bf 2.21} & \textbf{3.25}\\
        \bottomrule
    \end{tabularx}
    \vspace{-10pt}
\end{table}
\vspace{-5pt}

\subsection{Smear Estimation Ablation}
\label{subsec:smear_estim}
In this section, we demonstrate our contributions: using double-angle representation and the forward-backward cross consistency check in smear path learning.
We experiment with three different variants of representation for smear vector estimation: 1) Using local vectors used in~\cite{argaw2021optical} and learning them directly regardless of their vector-signs; 2) Learning with the restricted vectors used in~\cite{gong2017motion} that enforce positive horizontal components; and 3) Learning the proposed double-angle vectors described in~\cref{subsec:smear-est}.
We additionally evaluate the models without the involvement of the cross check map, with the training loss function in~\cref{eq:loss_no_cr} instead of~\cref{eq:loss}.
The evaluation results are reported in~\cref{tab:EPE-S_exp}, showing that using double angles consistently yields better results compared to prior work (local and restricted).
We argue that representing smear paths in double-angle space simplifies the the learning problem, by avoiding discontinuities.
In addition, the usage of the cross check map achieves significant improvements by avoiding training on unreliable blur sources in the image.
On average, using double-angle vectors and a cross-check map results in the performance gains of 40.26\% compared to restricted vectors and 52.90\% compared to local vectors.
\vspace{-0.5em}
\subsection{Uncertainty Ablation}
\label{subsec:ablation}
\begin{figure}[t]
    \centering
    \begin{subfigure}{.4\linewidth}
        \includegraphics[width=\linewidth]{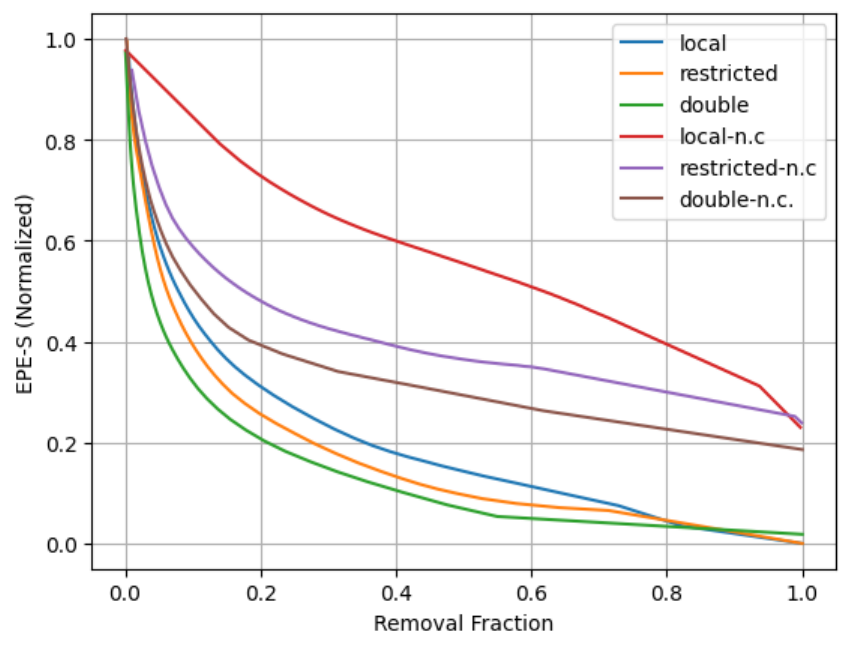}
        \subcaption{Smear prediction sparsification.}
        \label{fig:sub-fig-spar-plot}
    \end{subfigure} \hspace{1pt}%
    ~
    \begin{subfigure}{.4\linewidth}
        \includegraphics[width=\linewidth]{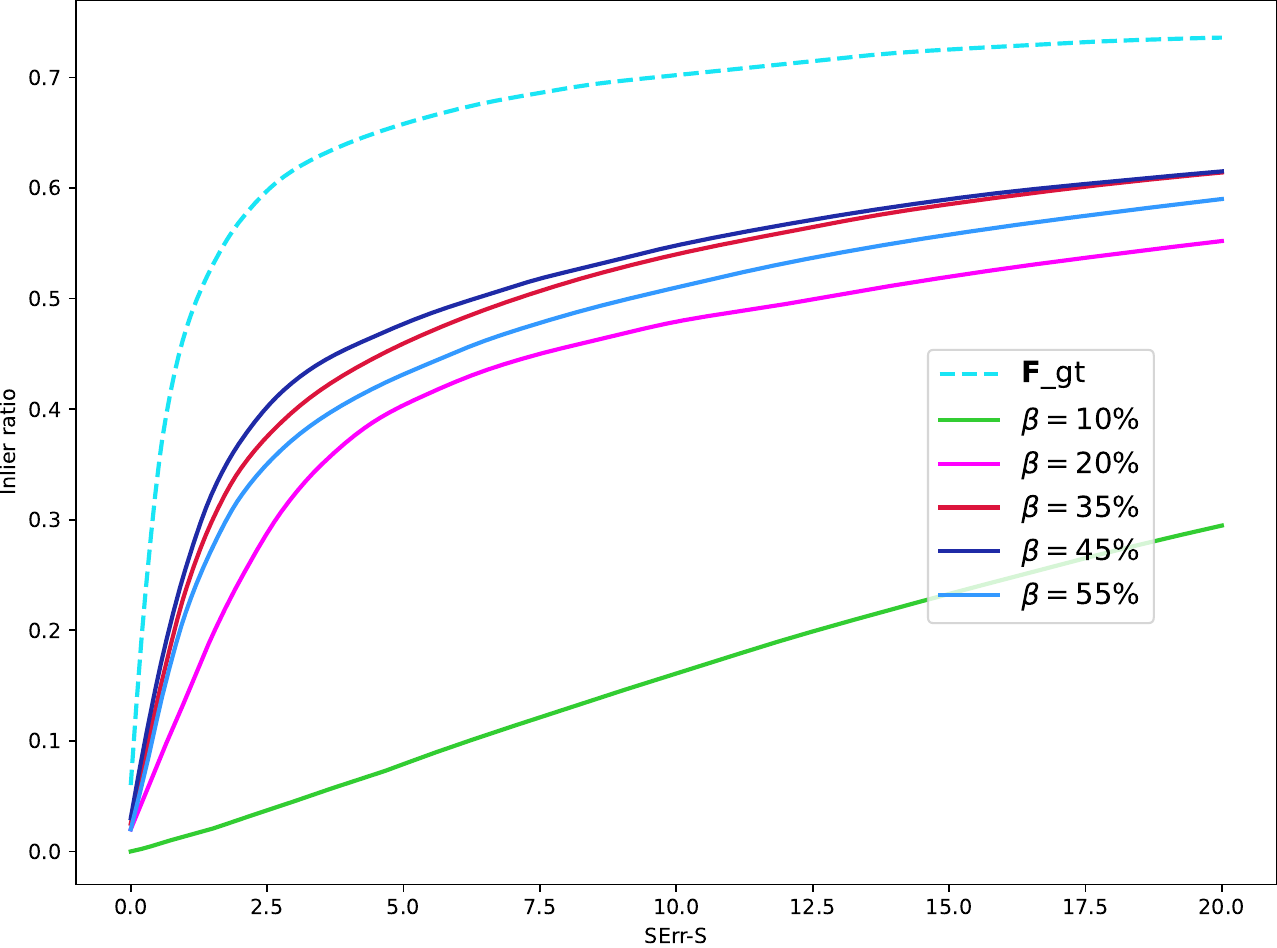}
        \subcaption{\textbf{F}-matrix cumulative errors.}
        \label{fig:sub-fig-cum-err}
    \end{subfigure}

    \caption{
    (a) Sparsification plots when using different smear representations on the Monkaa dataset.
    The ``-n.c.'' denotes the settings without the cross-check map, and lower EPE-S is better.
    (b) The CDE curves for the fundamental matrix estimation on the Monkaa dataset.  
    The dashed curve depicts the GT fundamental matrix, other curves correspond to different subset sizes $\mathcal{S}_\beta$, and closer to GT curve is better.
    \label{fig:ablation-fig}
    }
\end{figure}

We ablate the impact of the uncertainty measurement in the robustness of geometric estimation.
We first investigate how uncertainty guidance relates to smear accuracy in the sparsification plot in~\cref{fig:sub-fig-spar-plot},
wherein the $x$-axis shows the increasing removal fraction of high-variance predictions, 
and the $y$-axis reports the normalized EPE-S metric on the remaining predictions.
This plot, commonly used for evaluating uncertainty measures \cite{lind24,ilg2018uncertainty}, shows how well the predicted variance correlates with the removal fractions, \ie the accuracy of the remaining smear subsets $\mathcal{S}_\beta$ after removing different fractions of uncertain predictions.
The lower curve indicates better smear estimation quality across different certainty levels.
The green curve (double) in~\cref{fig:sub-fig-spar-plot} overall shows the lowest EPE-S across a range of removal fractions, demonstrating that the use of double-angle representation and the cross check map in our method effectively improves the uncertainty.
In the~\cref{fig:sub-fig-cum-err}, we conduct an ablation study on the sampling subset size $\mathcal{S}_\beta$, using the optimal settings suggested by~\cref{fig:sub-fig-spar-plot} on the Monkaa dataset.
The sampled subset $S_\beta$ represents the remaining smear traces after discarding those with high uncertainty.
The effect of different sizes is shown via the cumulative distribution error (CDE) plots in~\cref{fig:sub-fig-cum-err}, which indicate the ratio of ground-truth smears that agree with the estimated fundamental matrix over a range of \SErrMin thresholds.
A higher curve indicates that the estimated geometry better captures the underlying global camera transformation.
The sparsification and CDE plots in~\cref{fig:ablation-fig} show the quality trade-off between the smear and the geometry fundamental matrix estimations.

%
%
\begin{table}[t]
    \scriptsize 
    \setlength{\tabcolsep}{5pt}
    \caption{The ablation study on the impact of various sample subset sizes $\mathcal{S}_\beta$ for the robust estimation of the fundamental matrix, described in~\cref{subsec:epipolar-est}.}
    \label{tab:ablation_beta}
    \centering
    \begin{tabularx}{\linewidth}{@{\extracolsep{\fill}}lcccccccccc}
        \toprule
        \multirow{2}{*}{Sample size} & \multicolumn{2}{c}{Monkaa} & \multicolumn{2}{c}{Driving} & \multicolumn{2}{c}{GoPro} & \multicolumn{2}{c}{RBI} & Average \\
        & inliers$\uparrow$ & med$\downarrow$  & inliers$\uparrow$ & med$\downarrow$ & inliers$\uparrow$ & med$\downarrow$ & inliers$\uparrow$ & med$\downarrow$ & inliers$\uparrow$ \\ 
        \midrule
        $\beta=20\%$ & 29.64 & 3.71 & 24.63 & 8.64 & 28.41 & 4.10 & 20.71 & 5.18 & 25.85 \\
        $\beta=35\%$ & 42.25 & 1.71 & 30.11 & 4.98 & {\bf 30.21} & 3.89 & 26.76 & 6.58 & 32.33 \\
        $\beta=45\%$ & {\bf 45.02} & 2.99 & {\bf 32.35} & 5.11 & 24.34 & 7.28 & {\bf 34.62} & 3.34 & \textbf{34.08} \\
        $\beta=55\%$ & 32.55 & 4.58 & 28.86 & 10.80 & 24.12 & 6.46 & 18.27 & 6.34 & 25.95 \\
        $\beta=100\%$ & 1.11 & 32.64 & 2.14 & 30.61 & 2.11 & 26.14 & 2.25 & 29.41 & 1.90 \\
        \bottomrule
    \end{tabularx}
    \vspace{-10pt}
\end{table}

This observation is further shown in experiments across different datasets reported in~\cref{tab:ablation_beta}, demonstrating that selecting too few or too many estimated smear paths does not benefit the global estimation.
The study suggests that overly small subsets (lower than 20\%) contain more accurate smear predictions but fail to generalize to global motion, while overly large subsets (higher than 55\%) cover the image better but include noisier smear predictions, degrading matrix estimation.
Note that setting $\beta=100\%$ means that the uncertainty is not in-use.
It dramatically degrades performance, suggesting that uncertainty plays an important role in our method.
Empirical results show that setting $\beta=45\%$ yields the highest average performance across all datasets.
\cref{tab:ablation_beta} experiments are conducted on a \textit{subset} of test set, therefore, the $\beta = 45\%$ row in~\cref{tab:ablation_beta} and the GeoBlur row in~\cref{tab:fmatrix_inlier} are not expected to produce identical values.
\vspace{-0.5em}
\subsection{Motion Segmentation} 
\begin{figure}[t]
    \centering
    %
    \begin{subfigure}{.23\linewidth}
        \includegraphics[width=\linewidth]{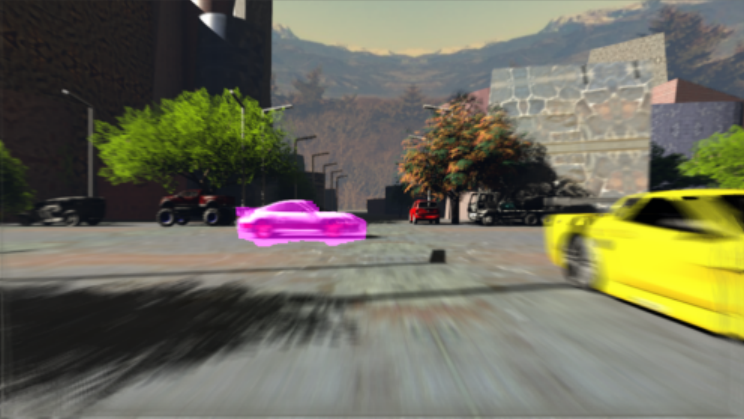}
    \end{subfigure}
    \begin{subfigure}{.23\linewidth}
        \includegraphics[width=\linewidth]{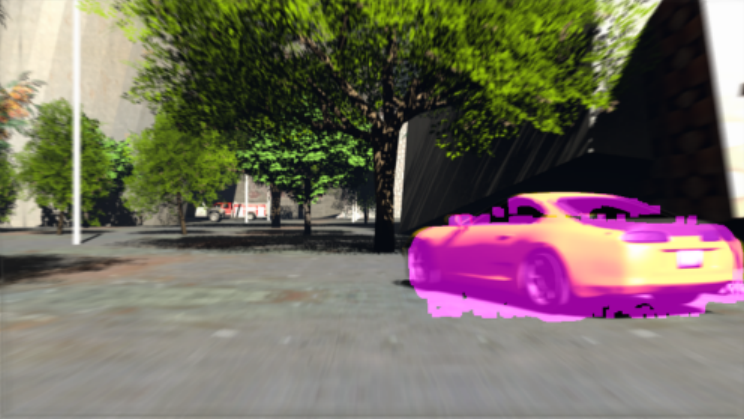}
    \end{subfigure}
    \begin{subfigure}{.23\linewidth}
        \includegraphics[width=\linewidth]{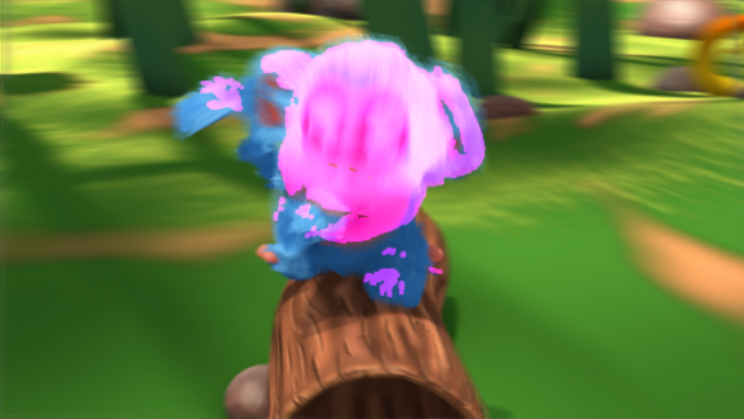}
    \end{subfigure}
    \caption{Example results on the motion segmentation task and the pink masks highlight objects exhibiting local motion.}
    \label{fig:motion_seg}
\end{figure}

We demonstrate our GeoBlur on the downstream task of motion segmentation, segmenting local moving objects that stand out from the global scene motion~\cite{argaw2021optical,Jain_2017_CVPR}.
We classify the estimated smear paths as compatible or incompatible with the estimated fundamental matrix, using thresholding in~\cref{eq:sampson_error}. 
The pixels that agree with the fundamental matrix are categorized as parts of global scene motion, and vice versa.
As shown in~\cref{fig:motion_seg}, moving objects, either partially or entirely, are classified as local motion based on the examined geometric consistency with global motion.
Our method can detect fast local motion, \eg a crossing car in the first image and a monkey moving head in the third image,
or a quick left-turning car in the second image.
In the third image, the local-classified region is also blurred, but this blur corresponds to local object motion rather than the global blur as in the background.

\vspace{-5pt}
\vspace{-0.3em}
\section{Conclusion}
\vspace{-0.5em}
We have proposed the method GeoBlur 
that estimates the fundamental matrix from a single blurred image, capturing the relative camera pose between shutter opening and closing.
Unlike conventional two-view methods, geometry estimation from a single blurred image is challenging due to missing correspondences and temporal ambiguity. 
Our approach predicts smear paths with per-pixel uncertainty and then robustly estimates the fundamental matrix from these.
It allows recovery of the underlying global camera motion while disregarding local object motion.
We show that, rather than discarding motion blur, smear traces provide valuable geometric cues for epipolar geometry estimation during camera exposure.
We further show that using the double-angle representation of smear paths and a cross-check map improves training effectiveness.
%

\noindent \textbf{Limitation and future work:}
In highly dynamic scenes, substantial independent motions may introduce many model outliers, making it more challenging to isolate the underlying camera motion. 
The future work could explore learning to identify local motions, which may require additional annotations not available in current datasets.

\clearpage  
%
%
\bibliographystyle{splncs04}
\bibliography{main}

@String(CVPR  = {IEEE Conf. Comput. Vis. Pattern Recog.})

@String(ICCV  = {Int. Conf. Comput. Vis.})

@String(ECCV  = {Eur. Conf. Comput. Vis.})

@String(NeurIPS = {Adv. Neural Inform. Process. Syst.})

@String(AAAI  = {AAAI})

@String(CVPR  = {CVPR})

@String(ICCV  = {ICCV})

@String(ECCV  = {ECCV})

@String(NeurIPS = {NeurIPS})

@book{hartley_zisserman2004,
    author      = {Hartley, Richard I. and Zisserman, Andrew},
    title       = {{Multiple View Geometry in Computer Vision}},
    address     = {{Cambridge, England, UK}},
    edition     = {2},
    isbn        = {0-521-54051-8},
    publisher   = {{Cambridge University Press}},
    year        = {2004}
}

@inproceedings{bigun87,
    author = {Josef Big\"un and Goesta H. Granlund},
    title = {Optimal Orientation Detection of Linear Symmetry},
    booktitle = {ICCV},
    year = {1987}
}

@book{sonkahlavacboyle2008,
  author = {Sonka, Milan and Hlavac, Vaclav and Boyle, Roger},
  title = {Image Processing: Analysis and Machine Vision},
  edition = 3,
  publisher = {Thomson Learning},
  year = 2008
}

@inproceedings{nister2004visual,
  title={Visual odometry},
  author={Nist{\'e}r, David and Naroditsky, Oleg and Bergen, James},
  booktitle={CVPR},
  OPTorganization={IEEE},
  year={2004},
}

@article{baker2011database,
  title={A database and evaluation methodology for optical flow},
  author={Baker, Simon and Scharstein, Daniel and Lewis, James P and Roth, Stefan and Black, Michael J and Szeliski, Richard},
  journal={International journal of computer vision},
  volume={92},
  number={1},
  pages={1--31},
  year={2011},
  publisher={Springer}
}

@inproceedings{brachmann18,
  title={Learning Less is More - {6D} Camera Localization via {3D} Surface Regression},
  author={Eric Brachmann and Carsten Rother},
  booktitle={CVPR},
  year={2018},
  organization={IEEE}
}

@inproceedings{pan2024glomap,
    author={Pan, Linfei and Barath, Daniel and Pollefeys, Marc and Sch\"{o}nberger, Johannes Lutz},
    title={Global Structure-from-Motion Revisited},
    booktitle={ECCV},
    year={2024}
}

@inproceedings{brooks2019learning,
  title={Learning to synthesize motion blur},
  author={Brooks, Tim and Barron, Jonathan T},
  booktitle={CVPR},
  year={2019},
    organization={IEEE}
}

@inproceedings{schonberger2016structure,
  title={Structure-from-motion revisited},
  author={Schonberger, Johannes L and Frahm, Jan-Michael},
  booktitle={CVPR},
  year={2016},
    organization={IEEE}
}

@inproceedings{rekleitis1996optical,
  title={Optical flow recognition from the power spectrum of a single blurred image},
  author={Rekleitis, Ioannis M},
  booktitle={3rd IEEE International Conference on Image Processing},
  year={1996},
  organization={IEEE}
}

@article{chen1996image,
  title={Image motion estimation from motion smear-a new computational model},
  author={Chen, Wei-Ge and Nandhakumar, N and Martin, Worthy N},
  journal={IEEE transactions on pattern analysis and machine intelligence},
  year={1996},
  publisher={IEEE}
}

@inproceedings{argaw2021optical,
  title={Optical flow estimation from a single motion-blurred image},
  author={Argaw, Dawit Mureja and Kim, Junsik and Rameau, Francois and Cho, Jae Won and Kweon, In So},
  booktitle={Proceedings of the AAAI conference on Artificial Intelligence},
  pages={891--900},
  year={2021}
}

@inproceedings{gong2017motion,
  title={From motion blur to motion flow: A deep learning solution for removing heterogeneous motion blur},
  author={Gong, Dong and Yang, Jie and Liu, Lingqiao and Zhang, Yanning and Reid, Ian and Shen, Chunhua and Van Den Hengel, Anton and Shi, Qinfeng},
  booktitle={CVPR},
  year={2017},
    organization={IEEE}
}

@inproceedings{mayer2016large,
  title={A large dataset to train convolutional networks for disparity, optical flow, and scene flow estimation},
  author={Mayer, Nikolaus and Ilg, Eddy and Hausser, Philip and Fischer, Philipp and Cremers, Daniel and Dosovitskiy, Alexey and Brox, Thomas},
  booktitle={CVPR},
  year={2016},
    organization={IEEE}
}

@inproceedings{nah2017deep,
  title={Deep multi-scale convolutional neural network for dynamic scene deblurring},
  author={Nah, Seungjun and Hyun Kim, Tae and Mu Lee, Kyoung},
  booktitle={CVPR},
  year={2017}
}

@inproceedings{purohit2019bringing,
  title={Bringing alive blurred moments},
  author={Purohit, Kuldeep and Shah, Anshul and Rajagopalan, AN},
  booktitle={CVPR},
  year={2019},
    organization={IEEE}
}

@inproceedings{jiang2018super,
  title={{Super SloMo}: High quality estimation of multiple intermediate frames for video interpolation},
  author={Jiang, Huaizu and Sun, Deqing and Jampani, Varun and Yang, Ming-Hsuan and Learned-Miller, Erik and Kautz, Jan},
  booktitle={CVPR},
  year={2018},
    organization={IEEE}
}

@inproceedings{menze2015object,
  title={Object scene flow for autonomous vehicles},
  author={Menze, Moritz and Geiger, Andreas},
  booktitle={CVPR},
  year={2015},
    organization={IEEE}
}

@article{luong1996fundamental,
  title={The fundamental matrix: Theory, algorithms, and stability analysis},
  author={Luong, Quan-Tuan and Faugeras, Olivier D},
  journal={International journal of computer vision},
  year={1996},
  publisher={Springer}
}

@inproceedings{wang2020learning,
  title={Learning feature descriptors using camera pose supervision},
  author={Wang, Qianqian and Zhou, Xiaowei and Hariharan, Bharath and Snavely, Noah},
  booktitle={ECCV},
  year={2020},
  organization={Springer}
}

@inproceedings{sun2018pwc,
  title={{PWC-Net}: {CNNs} for optical flow using pyramid, warping, and cost volume},
  author={Sun, Deqing and Yang, Xiaodong and Liu, Ming-Yu and Kautz, Jan},
  booktitle={CVPR},
  year={2018},
    organization={IEEE}
}

@inproceedings{dosovitskiy2015flownet,
  title={{FlowNet}: Learning optical flow with convolutional networks},
  author={Dosovitskiy, Alexey and Fischer, Philipp and Ilg, Eddy and Hausser, Philip and Hazirbas, Caner and Golkov, Vladimir and Van Der Smagt, Patrick and Cremers, Daniel and Brox, Thomas},
  booktitle={ICCV},
  year={2015}
}

@phdthesis{luong1992matrice,
  title={Matrice fondamentale et calibration visuelle sur l'environnement. Vers une plus grande autonomie des syst{\`e}me robotiques.},
  author={Luong, Quang-Tuan},
  year={1992},
  school={Universit{\'e} Paris Sud-Paris XI}
}

@article{huang1994motion,
  title={Motion and structure from feature correspondences: A review},
  author={Huang, Thomas S and Netravali, Arun N},
  journal={Proceedings of the IEEE},
    pages={252--268},
  year={1994}
}

@inproceedings{crandall2011discrete,
  title={Discrete-continuous optimization for large-scale structure from motion},
  author={Crandall, David and Owens, Andrew and Snavely, Noah and Huttenlocher, Dan},
  booktitle={CVPR},
  year={2011}
}

@inproceedings{jafarian2018monet,
  title={{MONET}: Multiview semi-supervised keypoint via epipolar divergence},
  author={Jafarian, Yasamin and Yao, Yuan and Park, Hyun Soo},
  booktitle = {ICCV},
  year={2019}
}

@article{zhang1998determining,
  title={Determining the epipolar geometry and its uncertainty: A review},
  author={Zhang, Zhengyou},
  journal={International journal of computer vision},
  volume={27},
  year={1998},
  publisher={Springer}
}

@inproceedings{wexler2003learning,
  title={Learning epipolar geometry from image sequences},
  author={Wexler, Yonatan and Fitzgibbon, Andrew W and Zisserman, Andrew},
  booktitle={CVPR},
  year={2003},
      organization={IEEE}
}

@inproceedings{ranftl2018deep,
  title={Deep fundamental matrix estimation},
  author={Ranftl, Ren{\'e} and Koltun, Vladlen},
  booktitle={ECCV},
  year={2018}
}

@article{lakshminarayanan2017simple,
  title={Simple and scalable predictive uncertainty estimation using deep ensembles},
  author={Lakshminarayanan, Balaji and Pritzel, Alexander and Blundell, Charles},
  journal={NeurIPS},
  volume={30},
  year={2017},
}

@article{longuet1981computer,
  title={A computer algorithm for reconstructing a scene from two projections},
  author={Longuet-Higgins, H Christopher},
  journal={Nature},
  year={1981},
  publisher={Nature Publishing Group UK London}
}

@article{nister2005preemptive,
  title={Preemptive {RANSAC} for live structure and motion estimation},
  author={Nist{\'e}r, David},
  journal={Machine Vision and Applications},
  year={2005},
  publisher={Springer}
}

@inproceedings{ilg2018uncertainty,
  title={Uncertainty estimates and multi-hypotheses networks for optical flow},
  author={Ilg, Eddy and Cicek, Ozgun and Galesso, Silvio and Klein, Aaron and Makansi, Osama and Hutter, Frank and Brox, Thomas},
  booktitle={ECCV},
  year={2018}
}

@article{sampson1982fitting,
  title={Fitting conic sections to “very scattered” data: An iterative refinement of the Bookstein algorithm},
  author={Sampson, Paul D},
  journal={Computer graphics and image processing},
  year={1982},
  publisher={Elsevier}
}

@inproceedings{fergus2006removing,
  title={Removing camera shake from a single photograph},
  author={Fergus, Rob and Singh, Barun and Hertzmann, Aaron and Roweis, Sam T and Freeman, William T},
  booktitle={SIGGRAPH},
  pages={787--794},
  year={2006}
}

@article{whyte2012non,
  title={Non-uniform deblurring for shaken images},
  author={Whyte, Oliver and Sivic, Josef and Zisserman, Andrew and Ponce, Jean},
  journal={International journal of computer vision},
  year={2012},
  publisher={Springer}
}

@inproceedings{fang2023self,
  title={Self-supervised non-uniform kernel estimation with flow-based motion prior for blind image deblurring},
  author={Fang, Zhenxuan and Wu, Fangfang and Dong, Weisheng and Li, Xin and Wu, Jinjian and Shi, Guangming},
  booktitle={CVPR},
  year={2023},
      organization={IEEE}
}

@inproceedings{kohler2012recording,
  title={Recording and playback of camera shake: Benchmarking blind deconvolution with a real-world database},
  author={K{\"o}hler, Rolf and Hirsch, Michael and Mohler, Betty and Sch{\"o}lkopf, Bernhard and Harmeling, Stefan},
  booktitle={ECCV},
  year={2012},
  organization={Springer}
}

@InProceedings{Jain_2017_CVPR,
author = {Dutt Jain, Suyog and Xiong, Bo and Grauman, Kristen},
title = {{FusionSeg}: Learning to Combine Motion and Appearance for Fully Automatic Segmentation of Generic Objects in Videos},
booktitle = {CVPR},
year = {2017},
    organization={IEEE}
}

@InProceedings{Zhong_2023_CVPR,
    author    = {Zhong, Zhihang and Cao, Mingdeng and Ji, Xiang and Zheng, Yinqiang and Sato, Imari},
    title     = {Blur Interpolation Transformer for Real-World Motion From Blur},
    booktitle = {CVPR},
    year      = {2023},
        organization={IEEE}
}

@InProceedings{Liang_2021_ICCV,
    author    = {Liang, Jingyun and Cao, Jiezhang and Sun, Guolei and Zhang, Kai and Van Gool, Luc and Timofte, Radu},
    title     = {{SwinIR}: Image Restoration Using Swin Transformer},
    booktitle = {ICCV Workshops},
    year      = {2021},
}

@Article{lind24,
  author = {Simon Kristofferson Lind and Ziliang Xiong and Per-Erik Forss\'en and Volker Kr\"uger},
  title = {Uncertainty Quantification Metrics for Deep Regression},
  journal = 	 {Pattern Recognition Letters},
  year = 	 {2024},
  volume = 	 {186},
  publisher =    {{Elsevier}},
}

@inproceedings{poursaeed2018deep,
  title={Deep fundamental matrix estimation without correspondences},
  author={Poursaeed, Omid and Yang, Guandao and Prakash, Aditya and Fang, Qiuren and Jiang, Hanqing and Hariharan, Bharath and Belongie, Serge},
  booktitle={ECCV Workshops},
  year={2018}
}

@inproceedings{ding2024fundamental,
  title={Fundamental matrix estimation using relative depths},
  author={Ding, Yaqing and V{\'a}vra, V{\'a}clav and Bhayani, Snehal and Wu, Qianliang and Yang, Jian and Kukelova, Zuzana},
  booktitle={ECCV},
    pages={142--159},
  year={2024},
}

@inproceedings{faugeras1992can,
  title={What can be seen in three dimensions with an uncalibrated stereo rig?},
  author={Faugeras, Olivier D},
  booktitle={ECCV},
  pages={563--578},
  year={1992},
  organization={Springer}
}

@inproceedings{chen2025imageimu,
  title={Image as an {IMU}: Estimating Camera Motion from a Single Motion-Blurred Image},
  author={Chen, Jerred and Clark, Ronald},
  booktitle={ICCV},
  year={2025}
}

@article{cvivsic2022soft2,
  title={{SOFT2}: Stereo visual odometry for road vehicles based on a point-to-epipolar-line metric},
  author={Cvi{\v{s}}i{\'c}, Igor and Markovi{\'c}, Ivan and Petrovi{\'c}, Ivan},
  journal={IEEE Transactions on Robotics},
  pages={273--288},
  year={2022},
  publisher={IEEE}
}

@article{Davison2007MonoSLAM,
  title={{MonoSLAM}: Real-Time Single Camera {SLAM}},
  author={Davison, Andrew J. and Reid, Ian D. and Molton, Nicholas D. and Stasse, Olivier},
  journal={IEEE Transactions on Pattern Analysis and Machine Intelligence},pages={1052--1067},
  year={2007}
}

@inproceedings{MurArtal2015MonocularVR,
  title={Monocular {SLAM} for User Viewpoint Tracking in Virtual Reality},
  author={Mur-Artal, Raul and Tardos, Juan D.},
  booktitle={IEEE International Conference on Robotics and Automation Workshops},
  year={2015}
}

@inproceedings{edstedt2024roma,
  title={{RoMa}: Robust dense feature matching},
  author={Edstedt, Johan and Sun, Qiyu and B{\"o}kman, Georg and Wadenb{\"a}ck, M{\aa}rten and Felsberg, Michael},
  booktitle={CVPR},
  pages={19790--19800},
  year={2024},
      organization={IEEE}
}

@inproceedings{wang2024dust3r,
  title={{DUSt3R}: Geometric {3D} vision made easy},
  author={Wang, Shuzhe and Leroy, Vincent and Cabon, Yohann and Chidlovskii, Boris and Revaud, Jerome},
  booktitle={CVPR},
  pages={20697--20709},
  year={2024},
      organization={IEEE}
}

@inproceedings{chum2005degensac,
  title     = {Two-view Geometry Estimation Unaffected by a Dominant Plane},
  author    = {Chum, Ond{\v{r}}ej and Werner, Tom{\'a}{\v{s}} and Matas, Ji{\v{r}}{\'i}},
  booktitle = {IEEE Conference on Computer Vision and Pattern Recognition},
  year      = {2005},
  pages     = {772--780},
}

\clearpage
\setcounter{page}{1}
\appendix

\section*{\centering Supplementary Materials}

This supplementary material includes five sections: (A) The comparison between two approaches to synthesize blur artifacts, to support why we use frame averaging method as in~\cref{subsec:smear-est}. 
We also show how the uncertainty map ground truth is acquired from forward-backward optical flow.
(B) More examples of our datasets, along with the smear-path predictions are shown in~\cref{fig:supp_smear_examples}. 
(C) A derived expression for conversion to and from the double-angle representation used to describe the smear paths shows explicit implementation during inference, which has been mentioned in~\cref{subsec:smear-est}. 
(D) Experiments on changing the value of the hyperparameter $\alpha$ in the loss function. 
(E) More quantitative demonstrations of fundamental matrix estimation, similar to those in~\cref{fig:qualitative_results}, are presented. 
Examples of challenging cases are provided in~\cref{fig:supp_fmatrix_failure_examples}.

\section{Comparison of blur generation methods}
\label{sec:comparison_with_psf_kernel}
\begin{figure}[ht]
    \centering
    \begin{subfigure}{.2776\linewidth}
        \includegraphics[width=\textwidth]{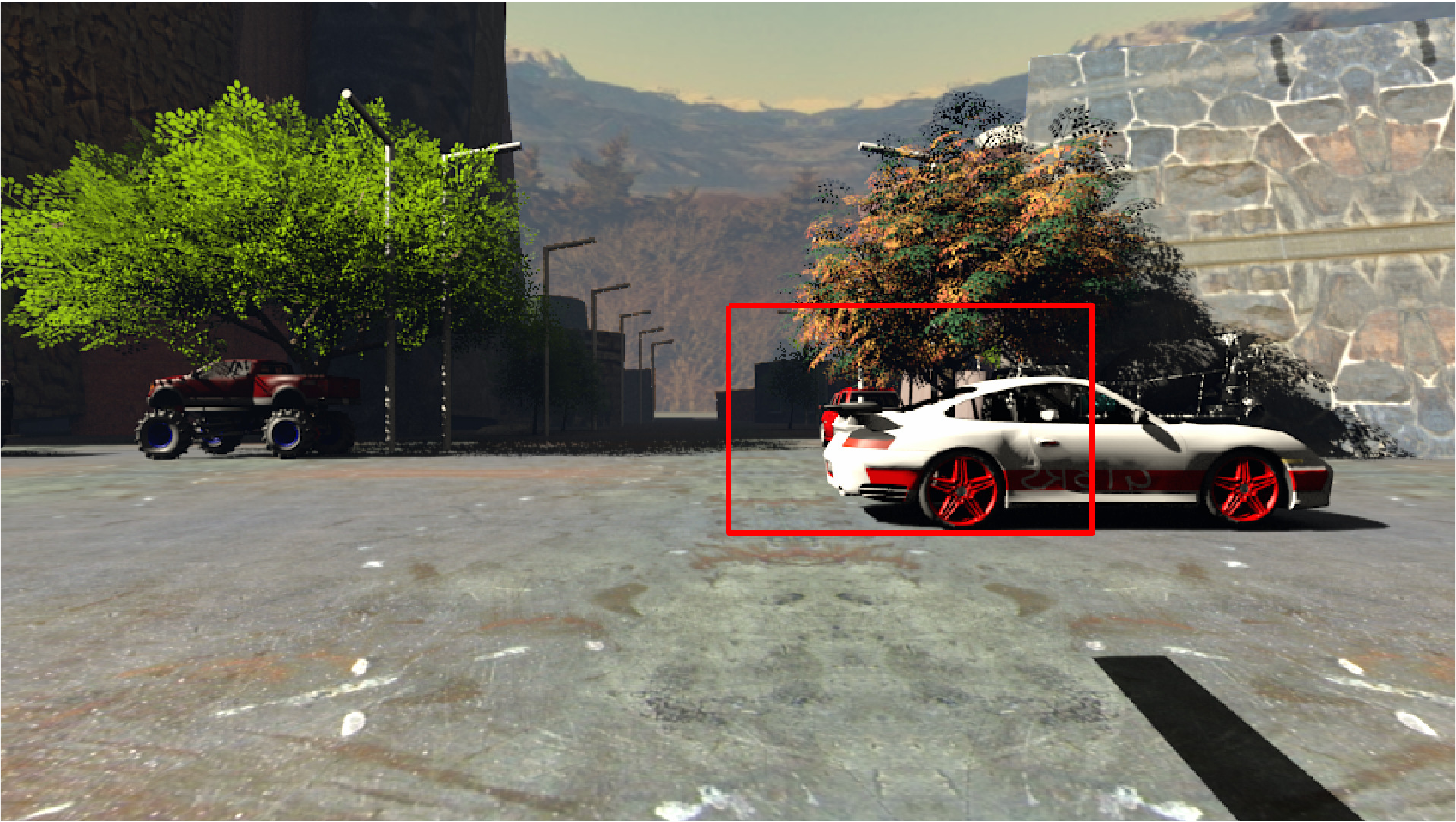}
    \end{subfigure}
    ~
    \begin{subfigure}{.25\linewidth}
        \includegraphics[width=\textwidth]{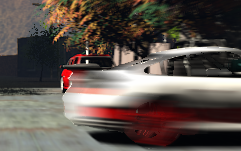}
    \end{subfigure}
    ~
    \begin{subfigure}{.25\linewidth}
        \includegraphics[width=\textwidth]{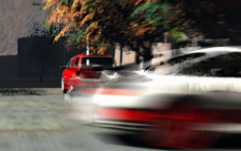}
    \end{subfigure}
    \caption{Comparison of synthetic blur using convolution and frame averaging. The car is moving to the right, and should be blurred accordingly. Left: middle sharp frame. Middle: result of convolution. Right, bottom: result of frame averaging.}
\label{fig:blur_comparison}
\end{figure}

A comparison of blur generation from convolution (using code from~\cite{gong2017motion}) and average-interpolation frames (suggested in \cite{argaw2021optical}) is shown in~\cref{fig:blur_comparison}.
Each method has its own characteristic artifacts. The convolution approach works well on smooth surfaces, but has artifacts at depth discontinuities. These artifacts arise from the lack of information about scene content occluded during exposure and from the absence of a prior on the blur kernel. As can be seen in~\cref{fig:blur_comparison} (middle), the result is an artificially sharp rear edge of the car in the blurred image.

The average-interpolation frame proposed in~\cite{argaw2021optical} also has some artifacts, see~\cref{fig:blur_comparison} (right). These artifacts stem from the use of a frame interpolation network to obtain intermediate frames. 
Rendering intermediate frames directly from a simulation engine would likely eliminate these artifacts.
Overall, we consider the approach proposed in~\cite{argaw2021optical}, which relies solely on rendered frames, to be better, however, it would require the creation of an entirely new dataset.

\subsection{Forward-Backward flow consistency check}
\label{subsec:supp_crosscheck}
\begin{figure*}[t]
\centering
    \begin{subfigure}{.24\linewidth} 
        \includegraphics[width=\textwidth]{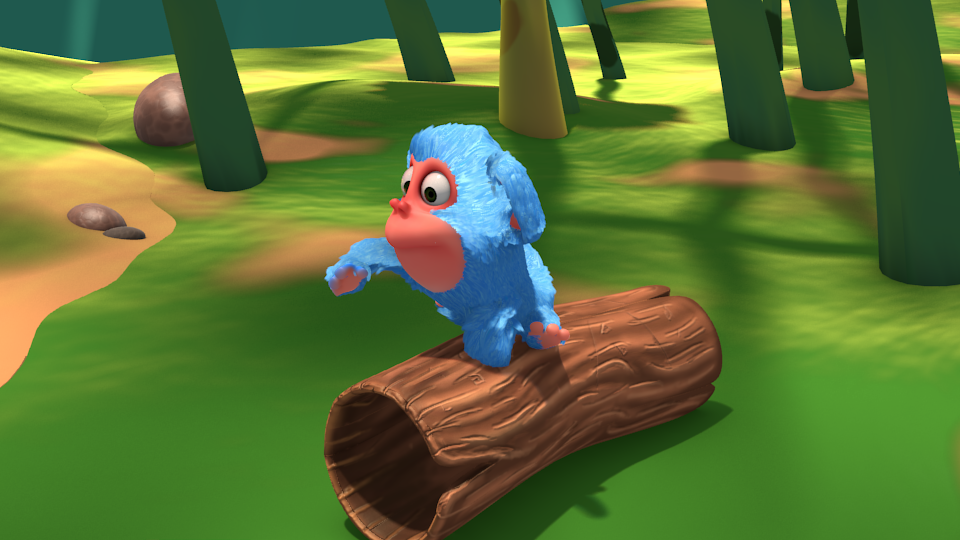}
        \subcaption{}
        \label{fig:fr_avg_a}
    \end{subfigure}
    \begin{subfigure}{.24\linewidth}
        \includegraphics[width=\textwidth]{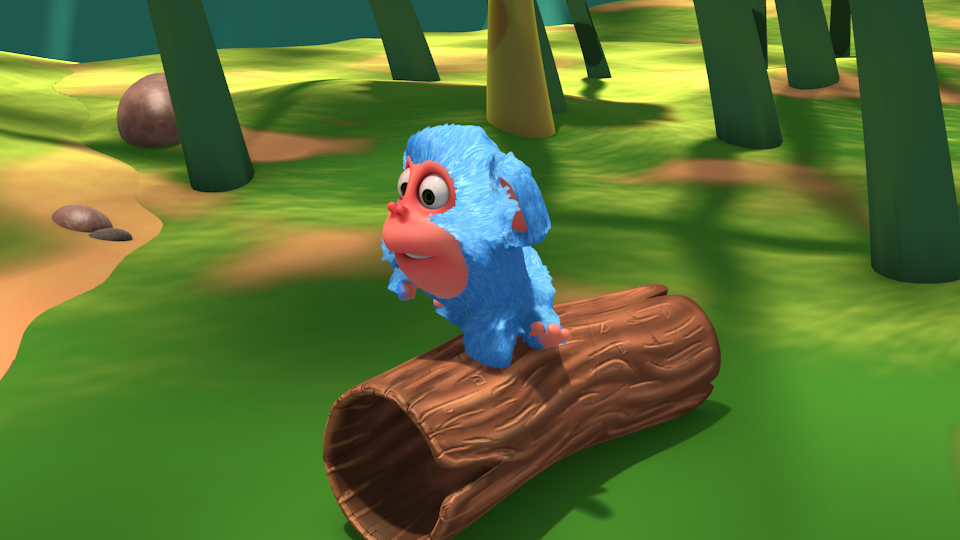}
        \subcaption{}
        \label{fig:fr_avg_c}
    \end{subfigure}
    \begin{subfigure}{.24\linewidth}
        \includegraphics[width=\textwidth]{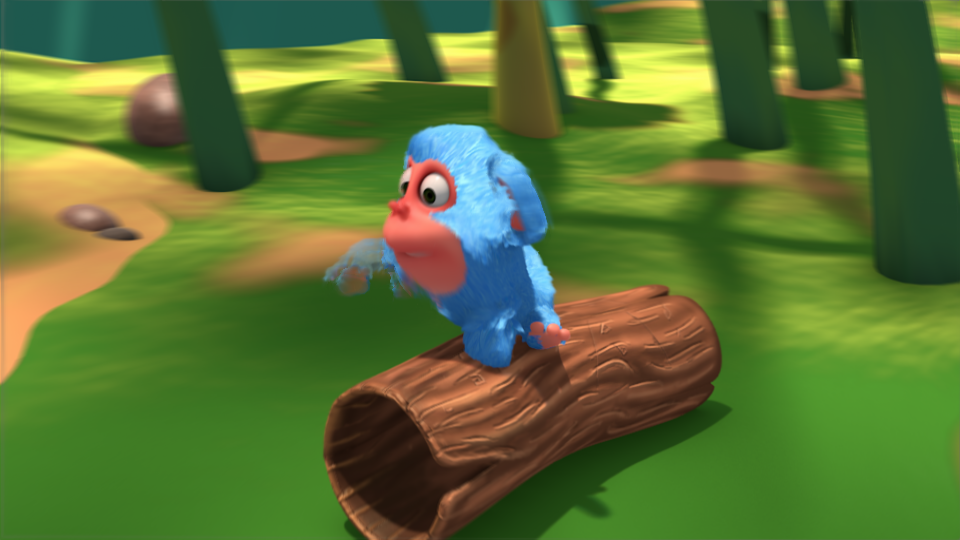}
        \subcaption{}
        \label{fig:fr_avg_d}
    \end{subfigure}
    \begin{subfigure}{.24\linewidth}
        \includegraphics[width=\textwidth]{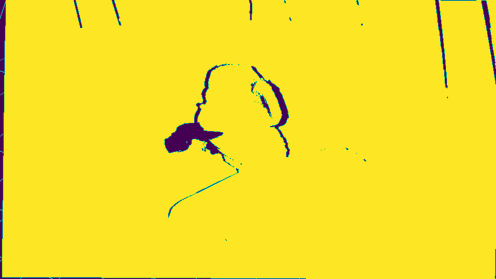}
        \subcaption{}
        \label{fig:fr_avg_e}
    \end{subfigure}
    \caption{Data generation example. (a), (b) Two consecutive sharp images. (c) The blurry image generated by averaging interpolated intermediate frames between two sharp images, as in \cite{argaw2021optical}. (d) The binary cross-check map indicating unreliable blurry patterns.}
    \label{fig:supp_fr_avg}
\end{figure*}

Here we provide an example of how the forward-backward flow cross-check is performed when producing smear annotations. 
This explains how the cross-check map can help the model avoid irregular smear paths during the learning procedure.
An example is shown in~\cref{fig:supp_fr_avg}, the object (the monkey) moves his arm largely in a very short time frame \ie from~\cref{fig:fr_avg_a} to~\cref{fig:fr_avg_c}. This causes missing interpolation information, leading to ghosting artifacts around the moving arm, as in~\cref{fig:fr_avg_d}. These unreliable blur artifacts are identified in the cross-check map shown in \cref{fig:fr_avg_e}.

\section{Implementation details}
\label{sec:optimization_detail.}
We provide implementation details for solving the optimization problem in~\cref{eq:fund_matrix_est} as follow.
At each RANSAC iteration, we randomly sample a 7-point subset $\mathcal{P}_7\subset\mathcal{P}_{\beta}$ and use~\cref{eq:fund_matrix_est} as a 7-point minimal-set estimator. 
We use \texttt{SciPy.\allowbreak optimize.\allowbreak minimize} with \texttt{trust-constr}, subject to $\|\mathbf{F}\|_F\!=\!1$ and $\det(\mathbf{F})\!=\!0$.
The optimization is initialized from a randomly generated $\mathbf{F}_0$, enforced to rank~2 and normalized to unit Frobenius norm.
To reduce computation, we apply an early \emph{preemptive rejection}: using $\mathbf{F}$  estimated from the \emph{algebraic} objective in~\cref{eq:fund_matrix_est}, we check its \emph{geometric} consistency on the same 7-point subset using the sign-agnostic Sampson error SErr-S in~\cref{eq:sampson}. 
If $\mathrm{SErr\text{-}S}>\tau_{\mathrm{SE7}}\!=\!1$, the hypothesis is discarded without full consensus evaluation.
Otherwise, inliers over $\mathcal{P}_{\beta}$ are counted using~\cref{eq:sampson} with $\tau_{\mathrm{SE}}=1$. RANSAC terminates after $1000$ iterations or when more than $80\%$ of $\mathcal{P}_{\beta}$ are inliers. 
These are summarized in the  pseudo-code in ~\cref{alg:preemptive_ransac}.

\captionsetup[algorithm]{font=small}
\begin{algorithm}[t]
\footnotesize
\caption{Preemptive RANSAC for Motion-blurred Geometry}
\label{alg:preemptive_ransac}
\begin{algorithmic}[1]
\Require Selected smear set $\mathcal{P}_{\beta}$

\State $\mathbf{F}^{*}\gets\varnothing,\;\mathcal{I}^{*}\gets\varnothing$
\For{$k=1,\ldots,N_{\max}=1000$}
    \State Sample $\mathcal{P}_{7}\subset\mathcal{P}_{\beta}$
    \State $\mathbf{F}\gets\arg\min_{\mathbf{F}}(\mathcal{P}_{7})$
    \Comment{Eq.~(9)}
    \If{$\mathrm{SErr\text{-}S}(\mathcal{P}_{7},\mathbf{F})
        >\tau_{\mathrm{SE7}}=1$}
        \State \textbf{continue}
        \Comment{Preemptive rejection}
    \EndIf
    \State $\mathcal{I}\gets
    \{p\in\mathcal{P}_{\beta}\mid
    \mathrm{SErr\text{-}S}(p,\mathbf{F})<\tau_{\mathrm{SE}}=1\}$
    \Comment{Eqs.~(10)--(11)}
    \If{$|\mathcal{I}|>|\mathcal{I}^{*}|$}
        \State $(\mathbf{F}^{*},\mathcal{I}^{*})
        \gets(\mathbf{F},\mathcal{I})$
    \EndIf
    \If{$|\mathcal{I}^{*}|/|\mathcal{P}_{\beta}|>\rho_{\mathrm{stop}}=0.8$}
        \State \textbf{break}
    \EndIf
\EndFor
\State \Return Optimal $\mathbf{F}^{*}$
\end{algorithmic}
\end{algorithm}

\section{Examples of datasets and smear predictions}
\label{sec:dataset}
\begin{figure}[t]
\centering
    \begin{subfigure}{.2\textwidth} 
        \includegraphics[width=\textwidth]{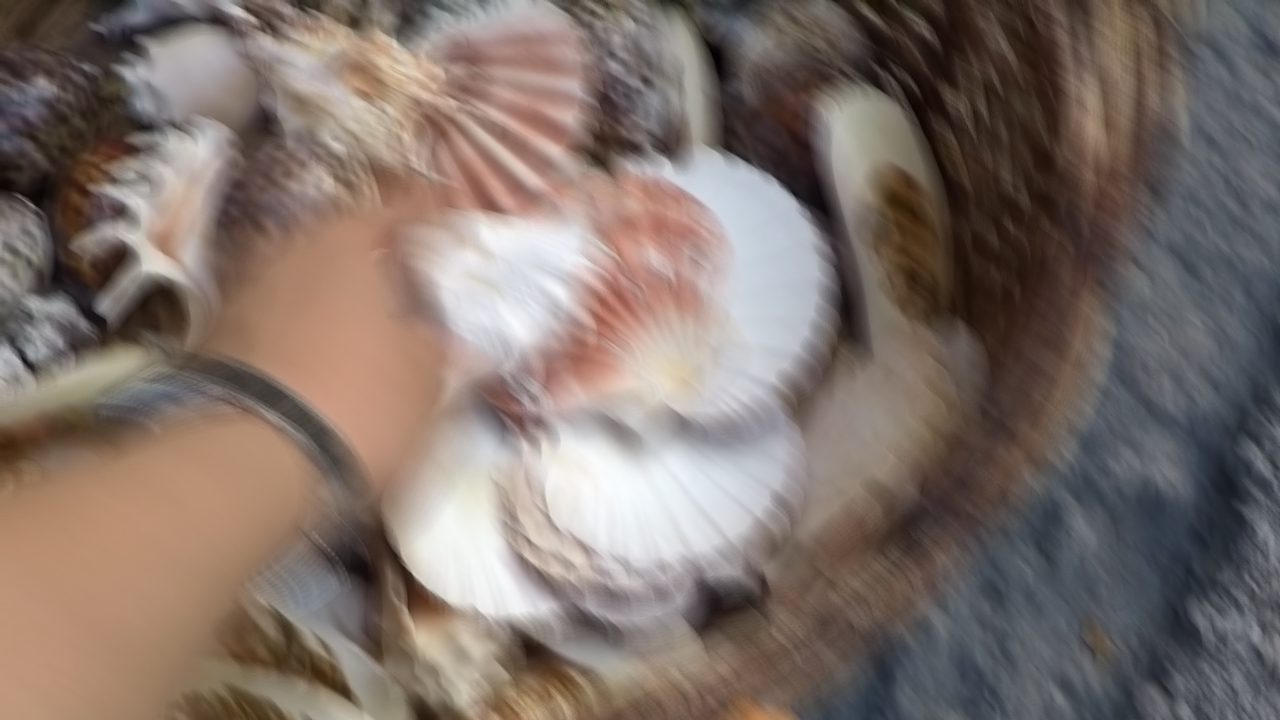}
    \end{subfigure}
    \begin{subfigure}{.2\textwidth}
        \includegraphics[width=\textwidth]{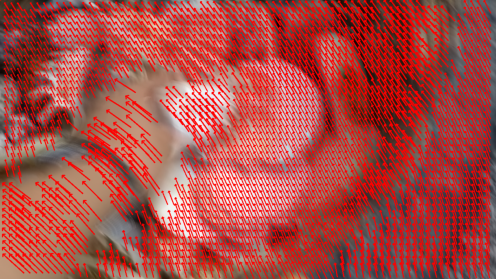}
    \end{subfigure}
    \begin{subfigure}{.2\textwidth}
        \includegraphics[width=\textwidth]{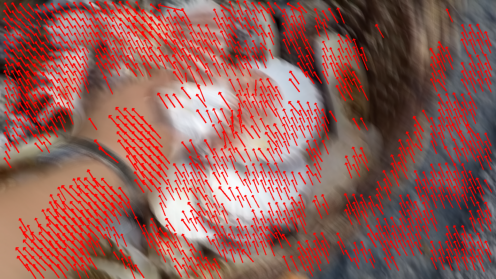}
    \end{subfigure}

    \begin{subfigure}{.2\textwidth} 
        \includegraphics[width=\textwidth]{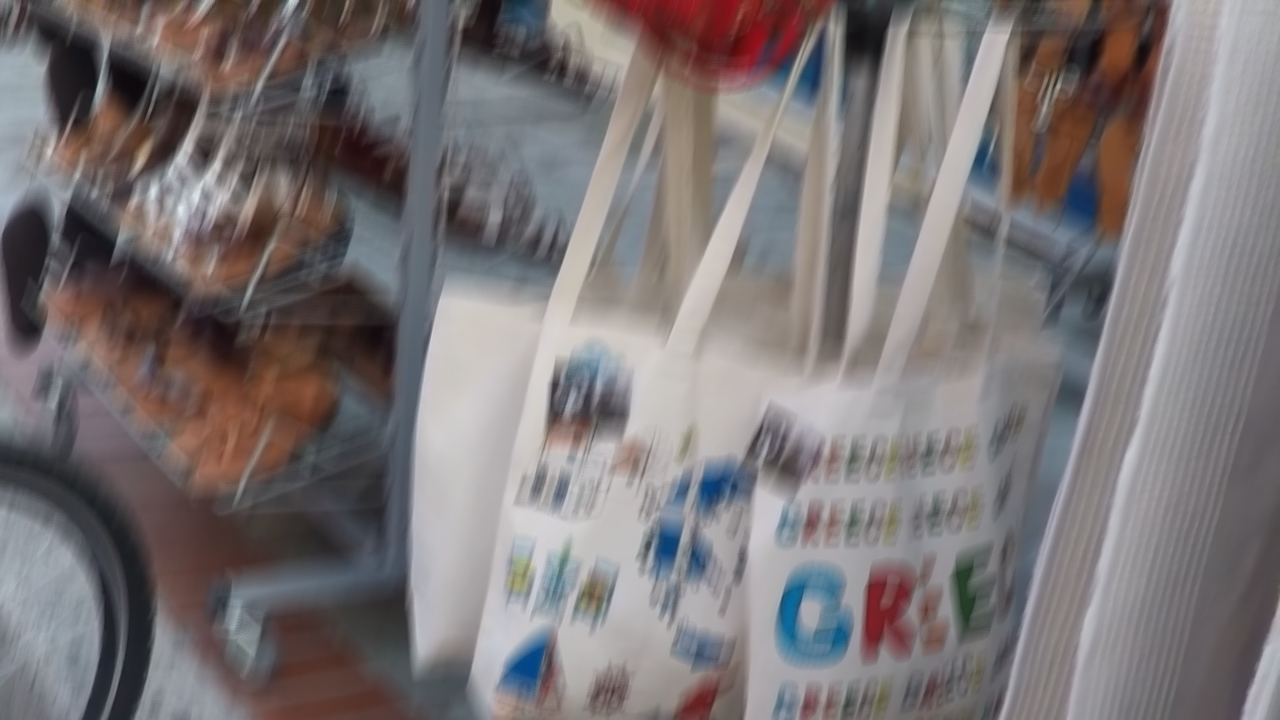}
    \end{subfigure}
    \begin{subfigure}{.2\textwidth}
        \includegraphics[width=\textwidth]{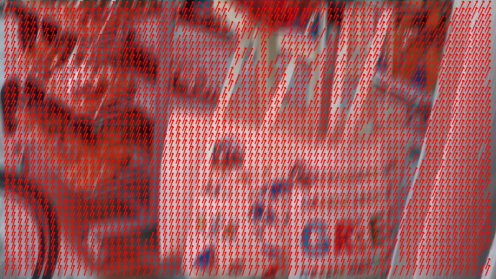}
    \end{subfigure}
    \begin{subfigure}{.2\textwidth}
        \includegraphics[width=\textwidth]{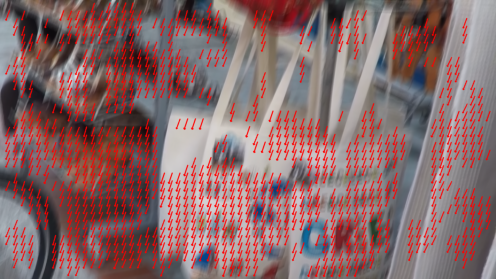}
    \end{subfigure}

    \begin{subfigure}{.2\textwidth} 
        \includegraphics[width=\textwidth]{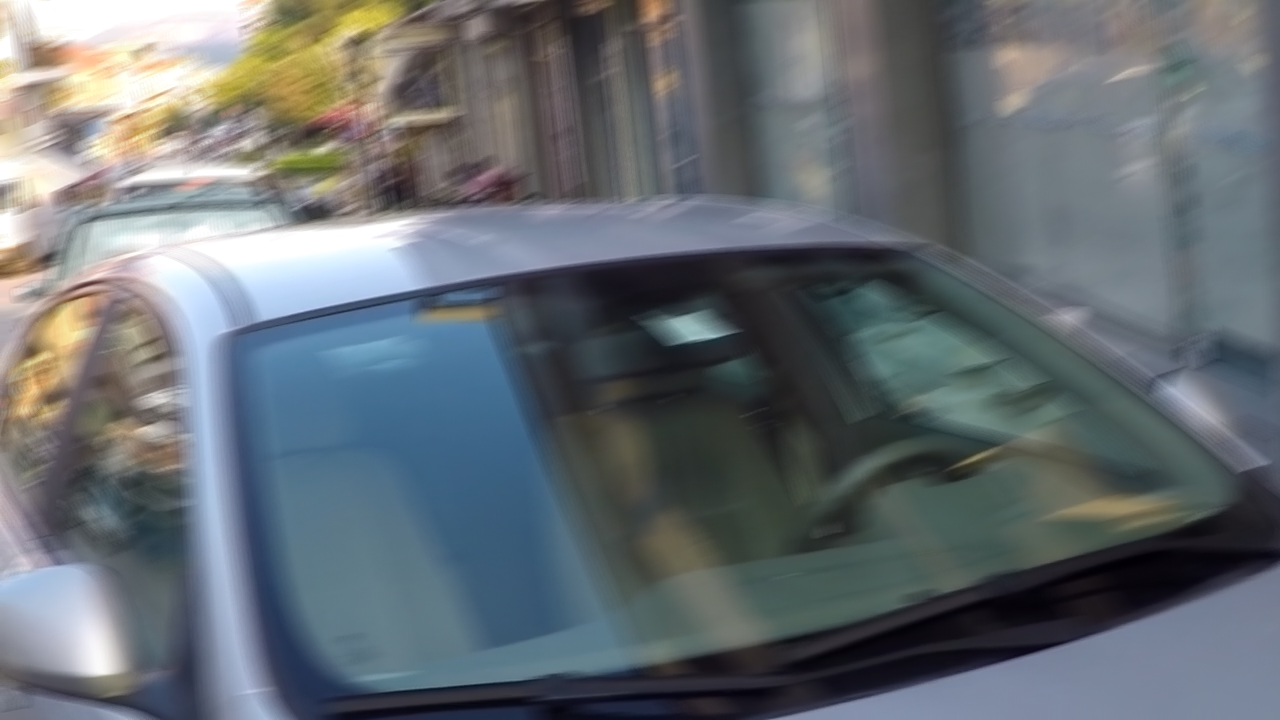}
    \end{subfigure}
    \begin{subfigure}{.2\textwidth}
        \includegraphics[width=\textwidth]{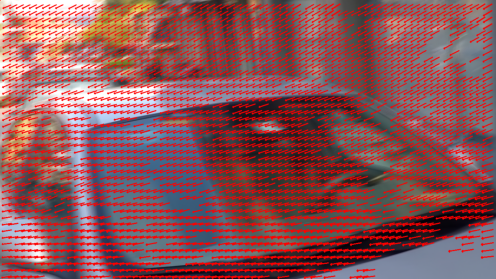}
    \end{subfigure}
    \begin{subfigure}{.2\textwidth}
        \includegraphics[width=\textwidth]{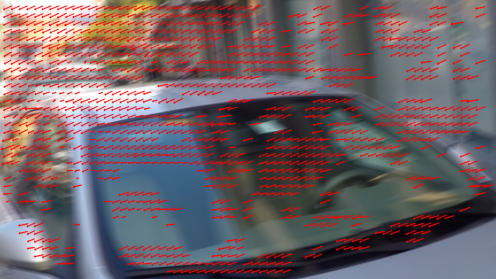}
    \end{subfigure}

    \begin{subfigure}{.2\textwidth} 
        \includegraphics[width=\textwidth]{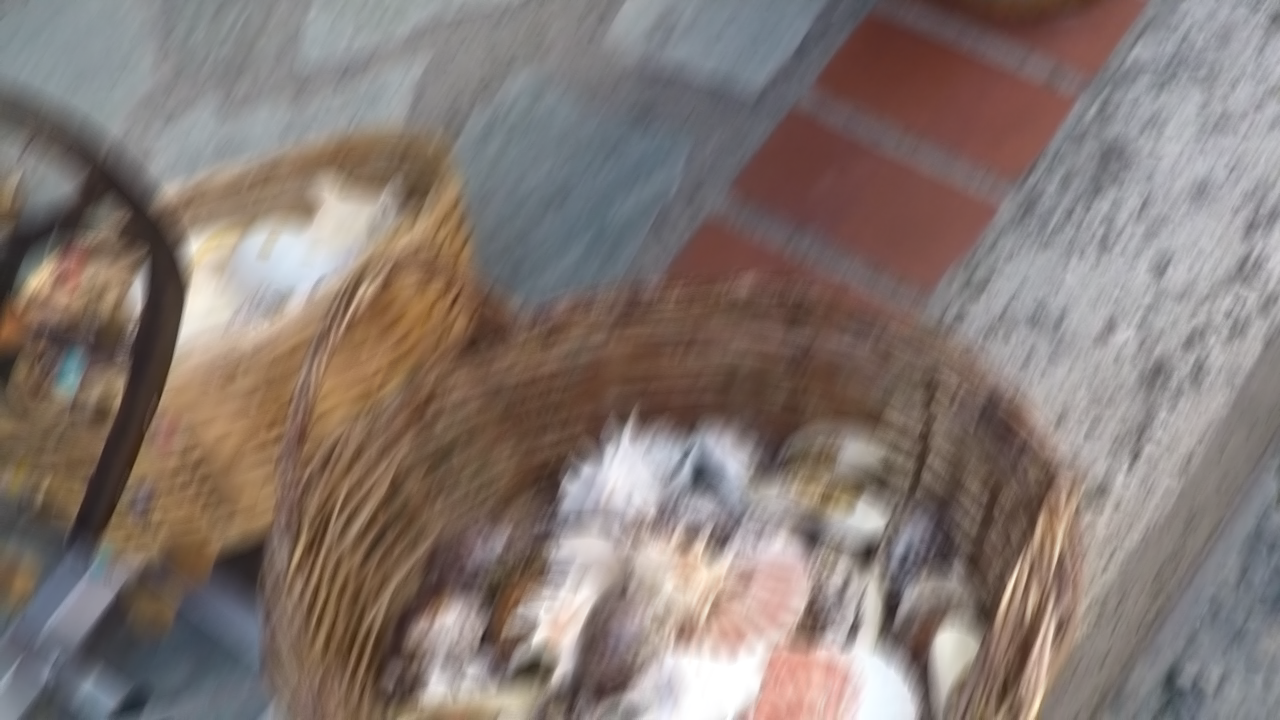}
    \end{subfigure}
    \begin{subfigure}{.2\textwidth}
        \includegraphics[width=\textwidth]{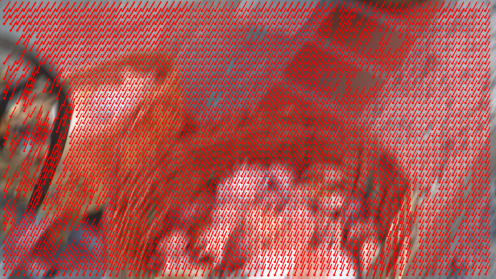}
    \end{subfigure}
    \begin{subfigure}{.2\textwidth}
        \includegraphics[width=\textwidth]{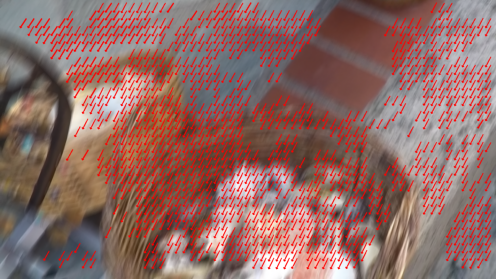}
    \end{subfigure}


    \begin{subfigure}{.2\textwidth} 
        \includegraphics[width=\textwidth]{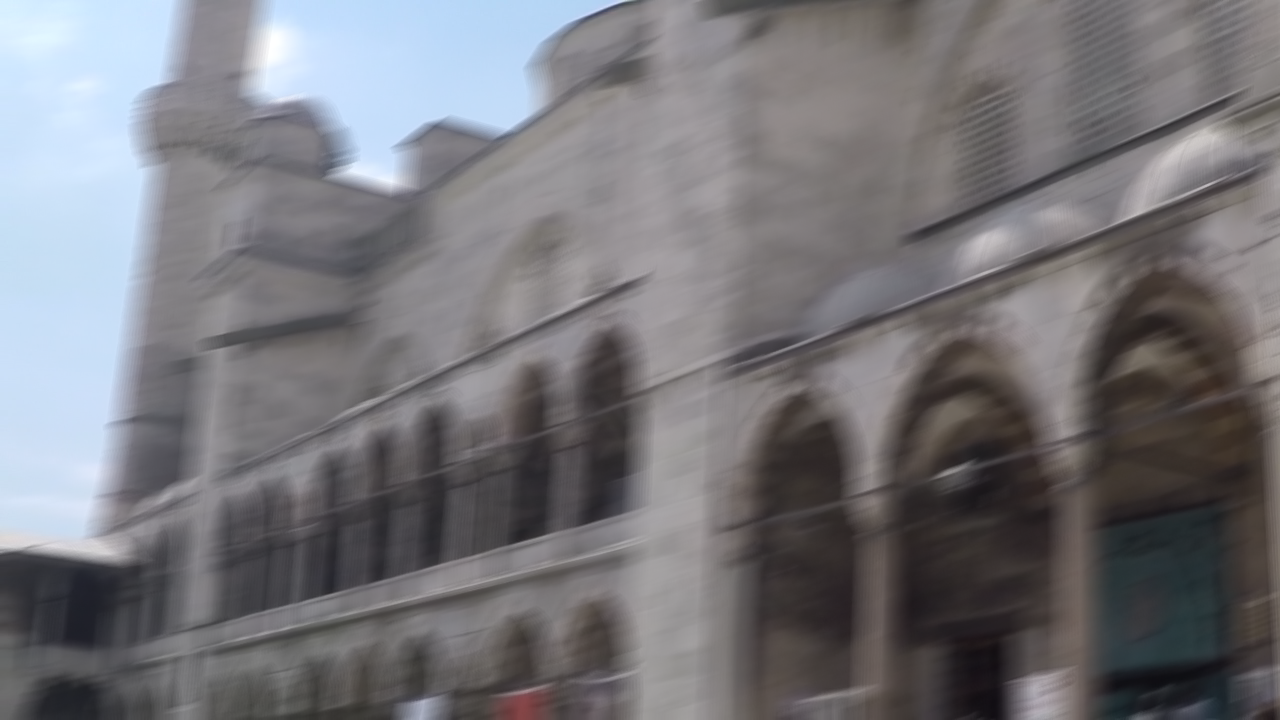}
    \end{subfigure}
    \begin{subfigure}{.2\textwidth}
        \includegraphics[width=\textwidth]{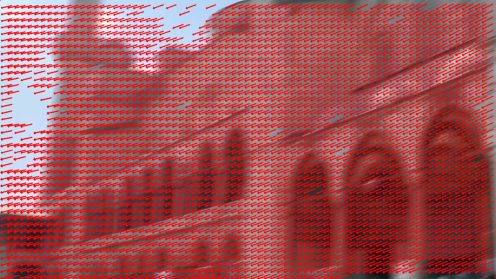}
    \end{subfigure}
    \begin{subfigure}{.2\textwidth}
        \includegraphics[width=\textwidth]{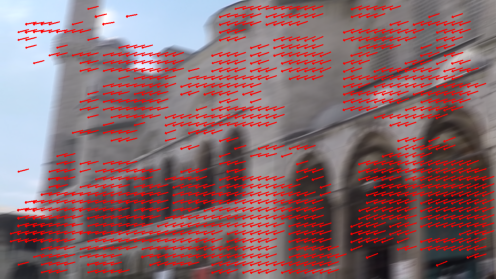}
    \end{subfigure}

    \begin{subfigure}{.2\textwidth} 
        \includegraphics[width=\textwidth]{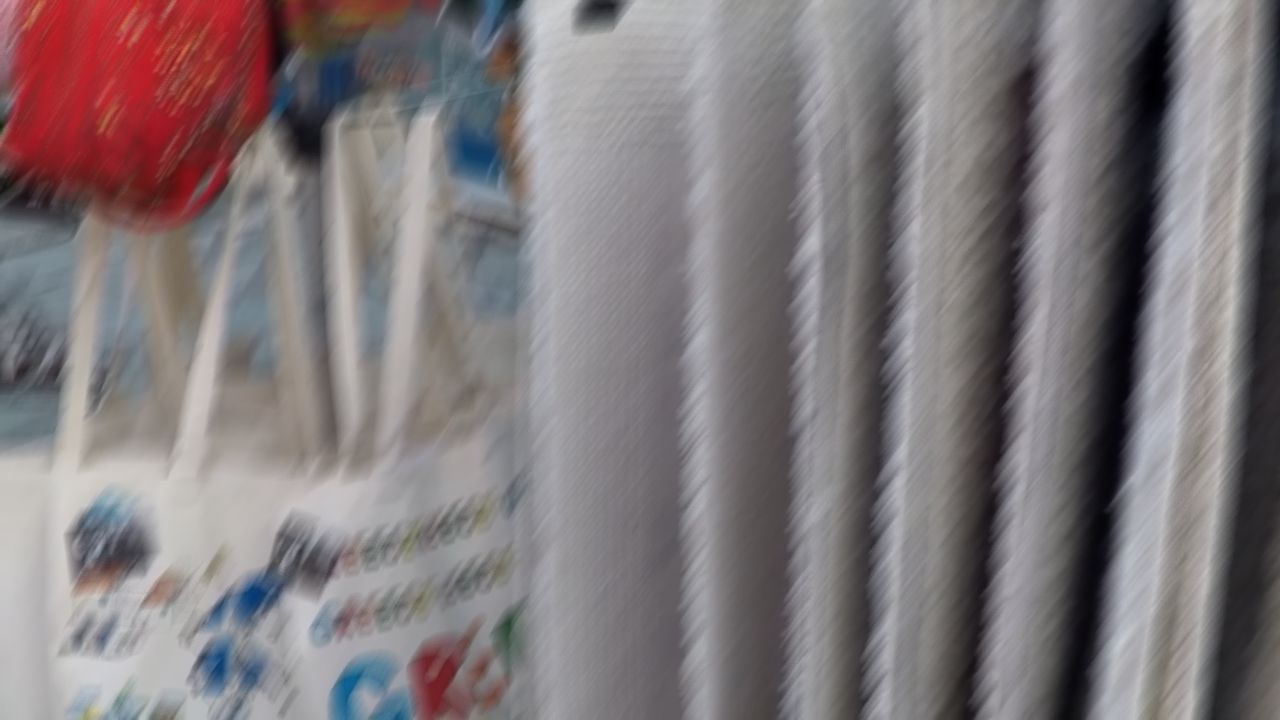}
    \end{subfigure}
    \begin{subfigure}{.2\textwidth}
        \includegraphics[width=\textwidth]{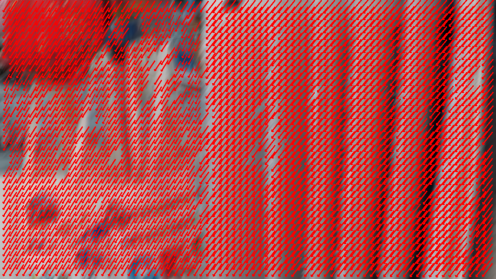}
    \end{subfigure}
    \begin{subfigure}{.2\textwidth}
        \includegraphics[width=\textwidth]{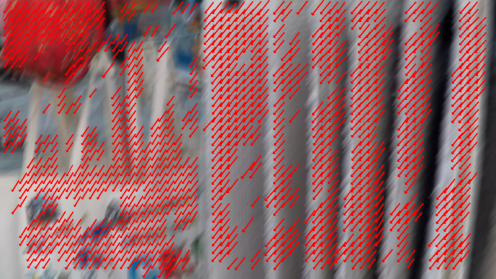}
    \end{subfigure}

    \begin{subfigure}{.2\textwidth} 
        \includegraphics[width=\textwidth]{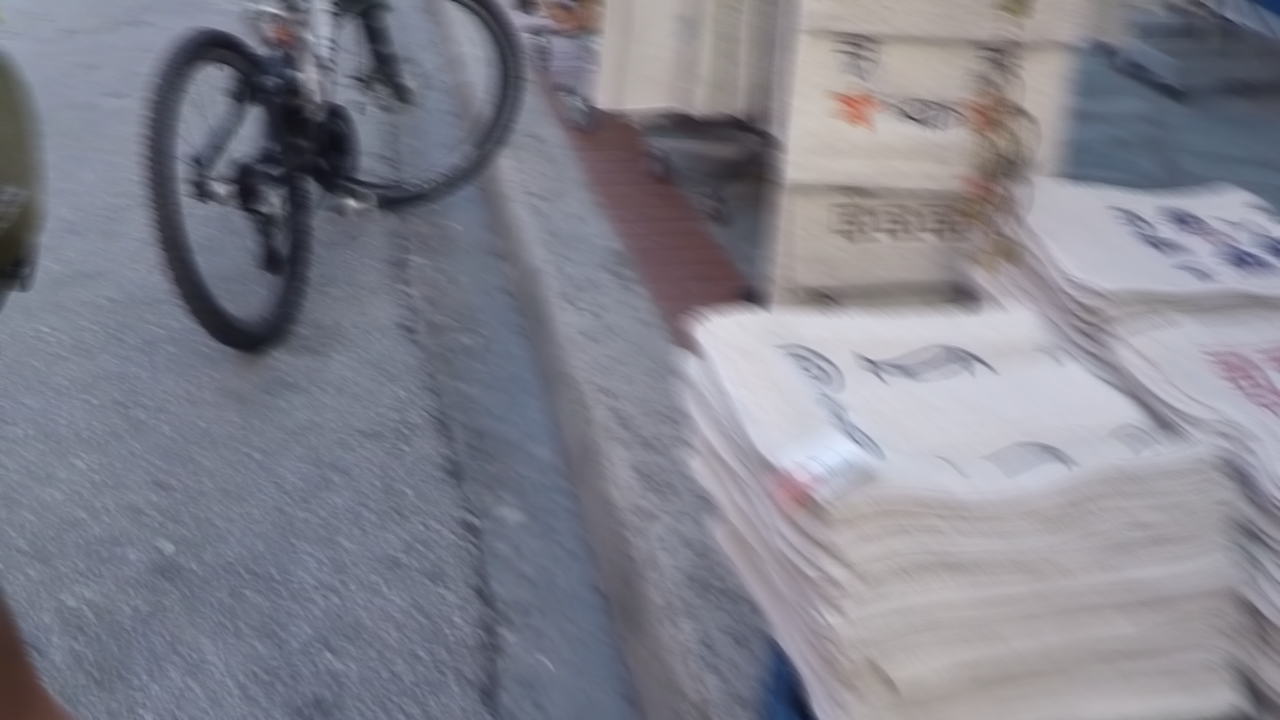}
    \end{subfigure}
    \begin{subfigure}{.2\textwidth}
        \includegraphics[width=\textwidth]{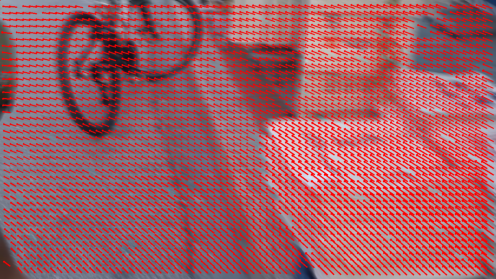}
    \end{subfigure}
    \begin{subfigure}{.2\textwidth}
        \includegraphics[width=\textwidth]{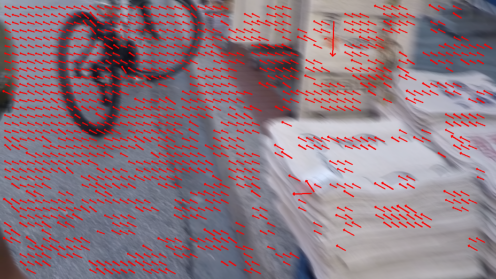}
    \end{subfigure}

    \caption{Examples of smear path estimation from the GoPro dataset. $1^{st}$ column is the input blurry image, $2^{nd}$ column is the generated ground truth smear paths, $3^{rd}$ column is the Top-50\% output of the smear-path prediction network, with column $1$ as input.}
    \label{fig:supp_smear_examples}
\end{figure}

This section provides examples of real scene datasets, and the results of smear path predictions, as seen Fig.\ \ref{fig:supp_smear_examples}. 
The $1^{st}$ and $2^{nd}$ columns show the motion-blurred images and the smear path annotations. The $3^{rd}$ column demonstrates Top-50\% best smear predictions ranked by the uncertainty measure $\sigma$. 
Accurate smear-path predictions are well distributed across the image, implying that geometry estimation from those reliably captures the global motion.
\vspace{-1em}

\section{Double angle conversion formulas}
\label{sec:supp_double_angle}
This section shows how we use double-angle representation to describe the smear path and how to convert it back to the local orientation during inference. The input is  a smear vector $s=(u,v)^\top$. As mentioned in the paper, this is inherently ambiguous, such that $s$ and $-s$ are equivalent. To derive the used double-angle expression, we first switch to polar coordinates
$d,\varphi$.

\begin{equation}
    s=\begin{pmatrix}u\\v\end{pmatrix}=d\begin{pmatrix}
        \cos \varphi \\
        \sin \varphi
    \end{pmatrix} \,.
    \label{eq:local}
\end{equation}
Now, using the double-angle formulas, we can formulate the smear path $s'$ as:

\begin{align}
    s'(p) &=\begin{pmatrix}
            u' \\
            v'
    \end{pmatrix}
        = d\begin{pmatrix}
        \cos 2\varphi \\
        \sin 2\varphi
    \end{pmatrix}
    = d\begin{pmatrix}
        \cos^2\varphi-\sin^2\varphi \\
        2\cos\varphi\sin\varphi
    \end{pmatrix} =
    \begin{pmatrix}
        \frac{u^2-v^2}{|s|} \\
        \frac{2uv}{|s|}
    \end{pmatrix}\,,
    \label{eq:double}
\end{align}
where $|s|=\sqrt{u^2+v^2}$.
From \eqref{eq:double}, we can see that the two local alternatives $(u,v)^\top$ and $(-u,-v)^\top$ are mapped to a single vector $(u',v')^\top$ (as only monomials in $u$, $v$ of degree 2 appear).
To go back from the double-angle representation, the smear length $d$ and local orientation $\varphi$ can be retrieved directly from $(u',v')^\top$:
\begin{equation}
    d=\sqrt{u'^2+v'^2}\quad\text{and}\quad \varphi=0.5\tan^{-1}(v',u')\,.
\end{equation}
Once $d$ and $\varphi$ are determined, $s=(u,v)^\top$ can be computed using \eqref{eq:local}.
\vspace{-5pt}

\section{Hyperparameter study on $\alpha$}

\label{sec:alpha_ablation}
\begin{table}[ht]
  \scriptsize
  \setlength{\tabcolsep}{7pt}
  \caption{Experiments with different $\alpha$ values.}
  \label{tab:ablation_alpha}
  \centering
  \begin{tabular}{lllll}
  \toprule
  \multirow{2}{*}{} & \multicolumn{3}{c}{EPE-S $\downarrow$} \\
         ~& Monkaa  & Driving & GoPro \\ 
    \hline
    $\alpha=$0.005 & 3.284 & 13.174 & 1.238\\
    $\alpha=$0.01 & {\bf 0.723} & {\bf 9.122} & {\bf 0.951} \\
    $\alpha=$0.02 & 1.387  & 18.370 & 1.457 \\ 
    $\alpha=$0.05 & 1.473 &  16.469 &  1.126 \\
    \bottomrule
  \end{tabular}
\end{table}

\begin{figure}[t]
\centering
    \begin{subfigure}{.21\linewidth}
        \includegraphics[width=\textwidth]{figs/Gopro_GOPR0372_07_01_000626.png}
    \end{subfigure}
    \begin{subfigure}{.21\linewidth}
        \includegraphics[width=\textwidth]{figs/first_Gopro_pre_pGT_626.png}
    \end{subfigure}
    \begin{subfigure}{.21\linewidth}
        \includegraphics[width=\textwidth]{figs/last_Gopro_pre_pGT_626.png}
    \end{subfigure}

    \begin{subfigure}{.21\linewidth}
        \includegraphics[width=\textwidth]{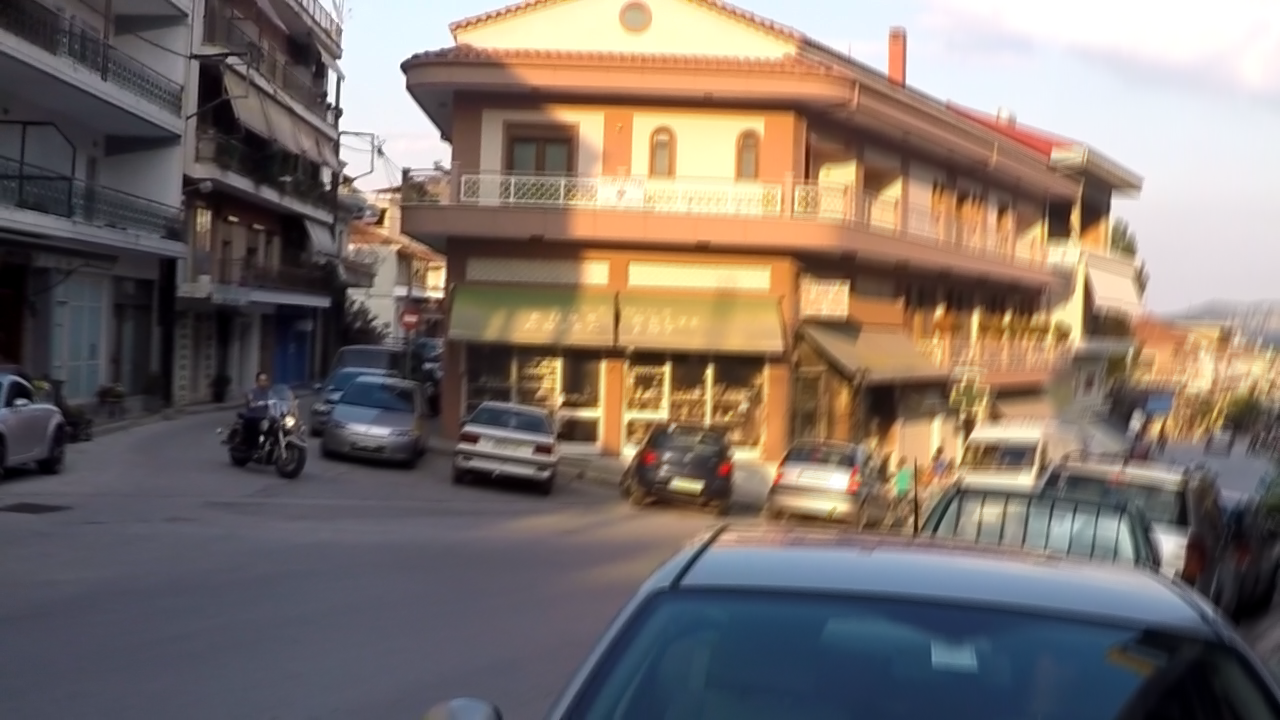}
    \end{subfigure}
    \begin{subfigure}{.21\linewidth}
        \includegraphics[width=\textwidth]{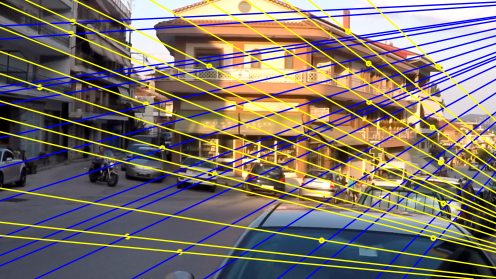}
    \end{subfigure}
    \begin{subfigure}{.21\linewidth}
        \includegraphics[width=\textwidth]{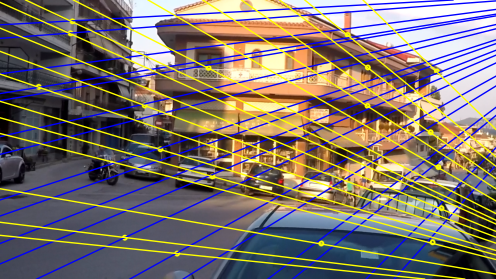}
    \end{subfigure}
    
    \begin{subfigure}{.21\linewidth} 
        \includegraphics[width=\textwidth]{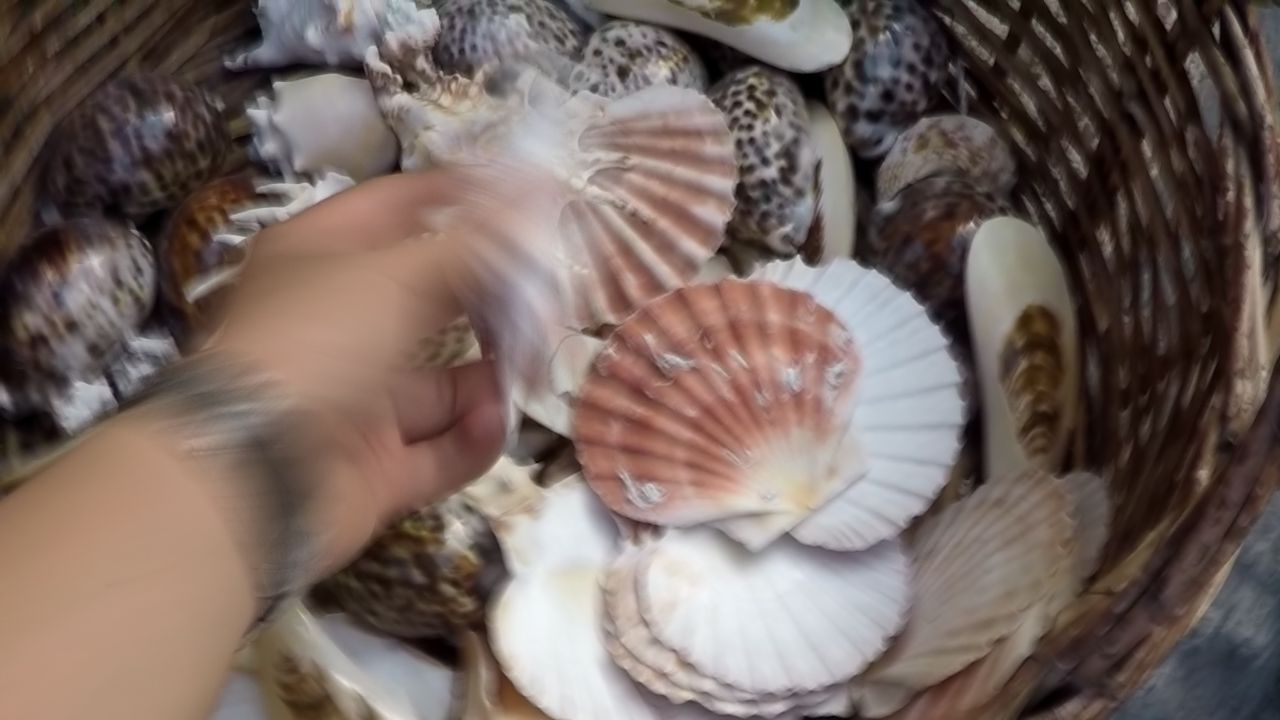}
    \end{subfigure}
    \begin{subfigure}{.21\linewidth}
        \includegraphics[width=\textwidth]{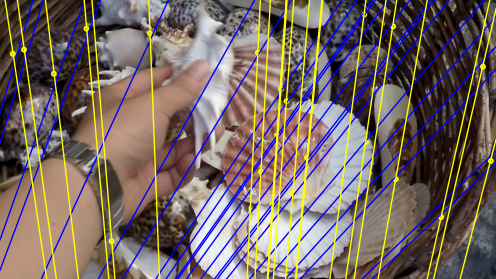}
    \end{subfigure}
    \begin{subfigure}{.21\linewidth}
        \includegraphics[width=\textwidth]{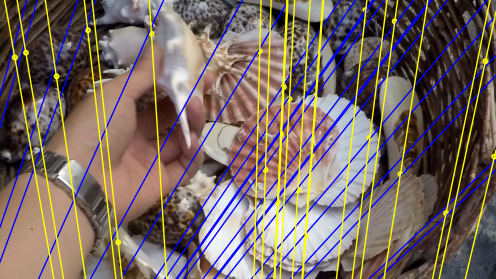}
    \end{subfigure}


    \begin{subfigure}{.21\linewidth} 
        \includegraphics[width=\textwidth]{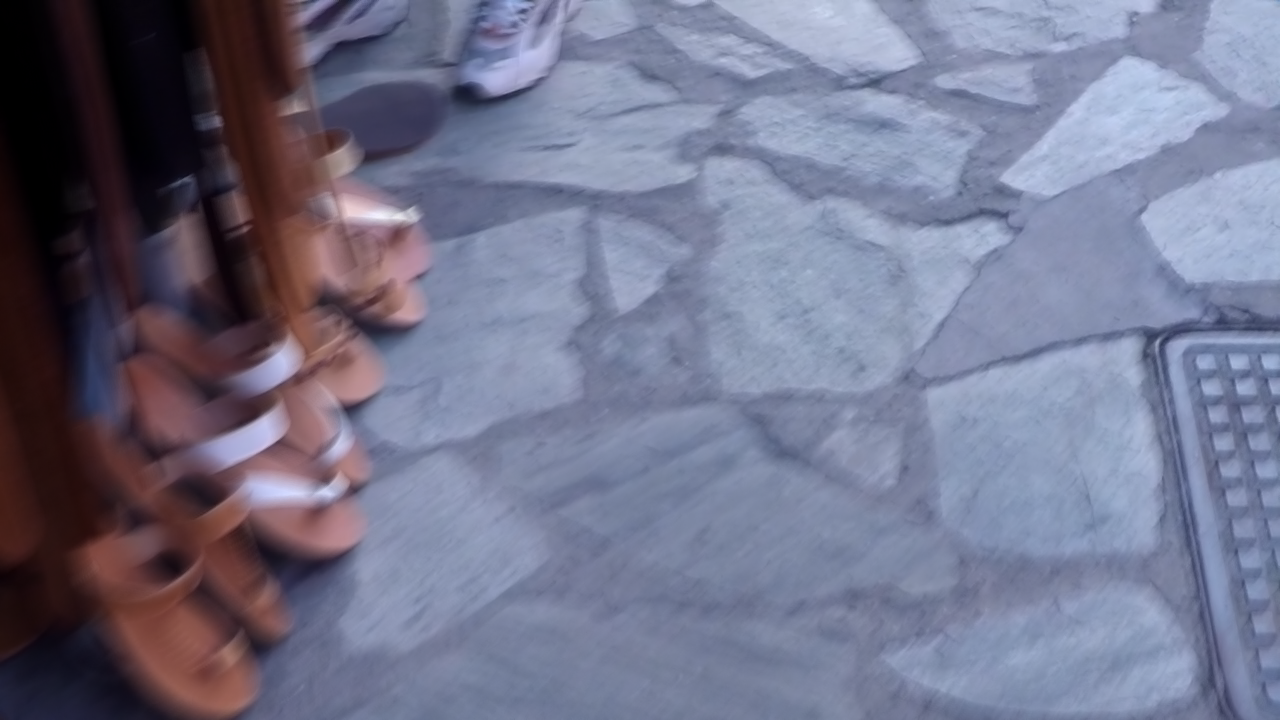}
    \end{subfigure}
    \begin{subfigure}{.21\linewidth}
        \includegraphics[width=\textwidth]{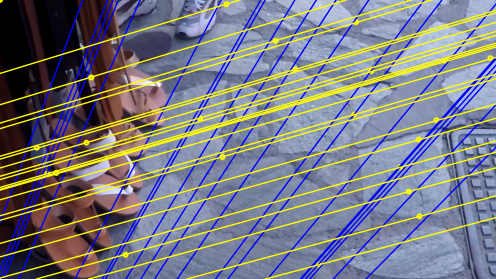}
    \end{subfigure}
    \begin{subfigure}{.21\linewidth}
        \includegraphics[width=\textwidth]{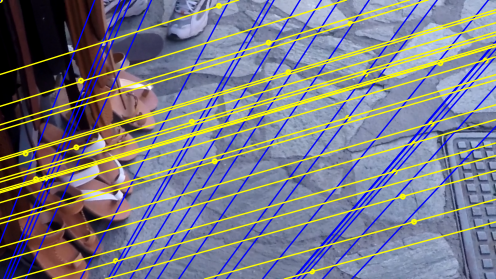}
    \end{subfigure}

    \begin{subfigure}{.21\linewidth} 
        \includegraphics[width=\textwidth]{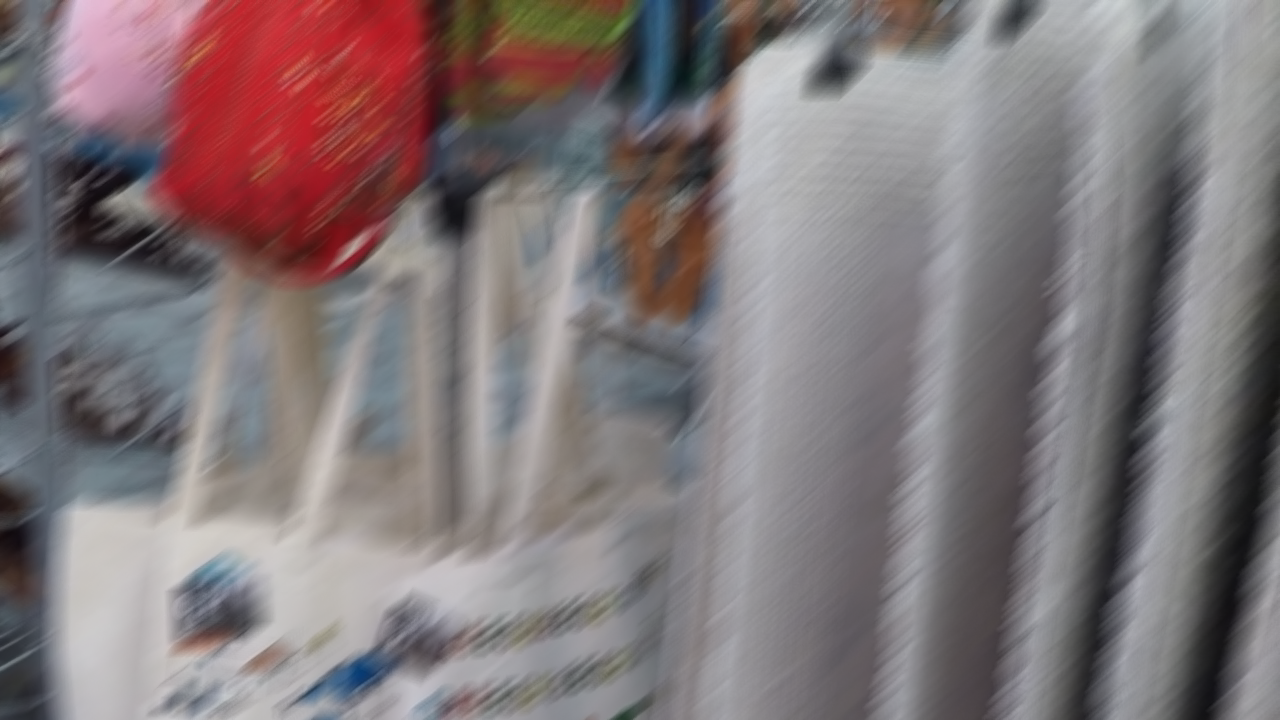}
    \end{subfigure}
    \begin{subfigure}{.21\linewidth}
        \includegraphics[width=\textwidth]{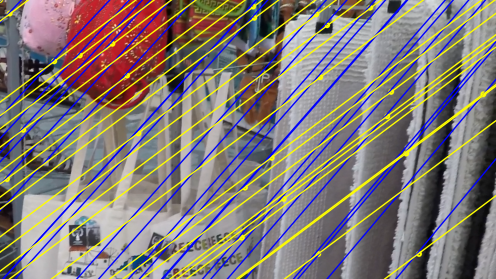}
    \end{subfigure}
    \begin{subfigure}{.21\linewidth}
        \includegraphics[width=\textwidth]{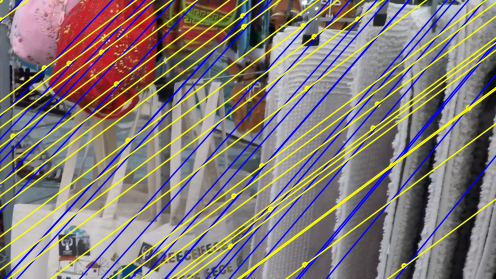}
    \end{subfigure}

    \begin{subfigure}{.21\linewidth} 
        \includegraphics[width=\textwidth]{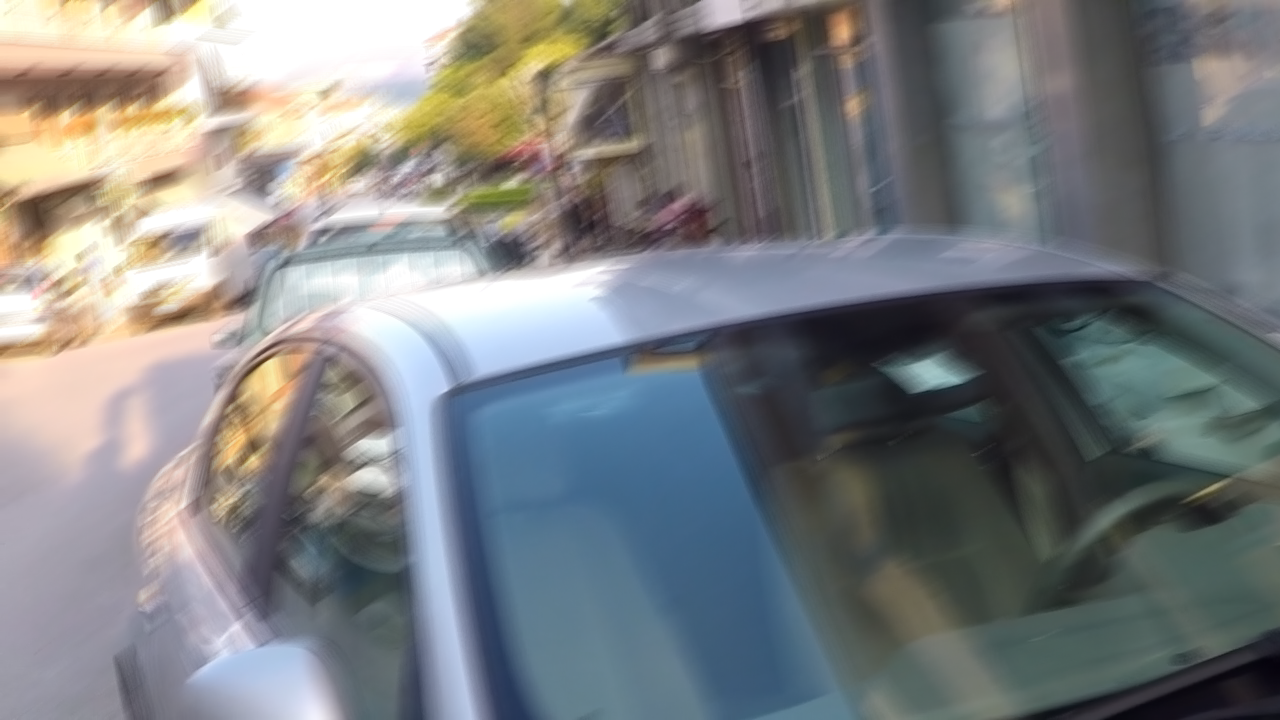}
    \end{subfigure}
    \begin{subfigure}{.21\linewidth}
        \includegraphics[width=\textwidth]{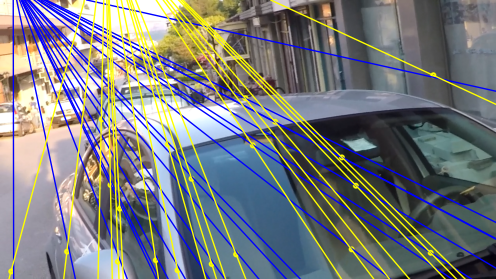}
    \end{subfigure}
    \begin{subfigure}{.21\linewidth}
        \includegraphics[width=\textwidth]{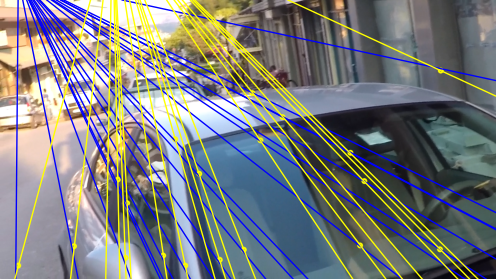}
    \end{subfigure}

    \caption{Additional quantitative examples from the GoPro dataset, similar to Fig. \ref{fig:qualitative_results}. First column: the blurry inputs. Second and third columns: the first and last frames, and corresponding epipolar lines.}
    \label{fig:supp_fmatrix_examples}
\end{figure}

In this section, we report experiments where we vary the hyperparameter $\alpha$ in the loss function \eqref{eq:loss}. The $\alpha$ parameter is introduced to prevent the uncertainty measure at the invalid pixels, identified by a value of $0$ in $m^{cr}$, from diverging to infinity. Empirical experiments from Tab.\ \ref{tab:ablation_alpha} suggest that $\alpha=0.01$ results in the lowest average end-point error, in all three datasets.

\section{Qualitative results and challenging cases}
\label{sec:qualitative_extra}
\vspace{-5pt}
Additional qualitative examples of our fundamental matrix estimation on realistic scenes are shown in Fig.\ \ref{fig:supp_fmatrix_examples}. As can be seen, our fundamental matrix results are close to the pseudo ground truth, which is derived from ground truth correspondences by the {\it gold standard} method \cite{hartley_zisserman2004}. 

\begin{figure}[t]
\centering
    \begin{subfigure}{.2\linewidth}
        \includegraphics[width=\textwidth]{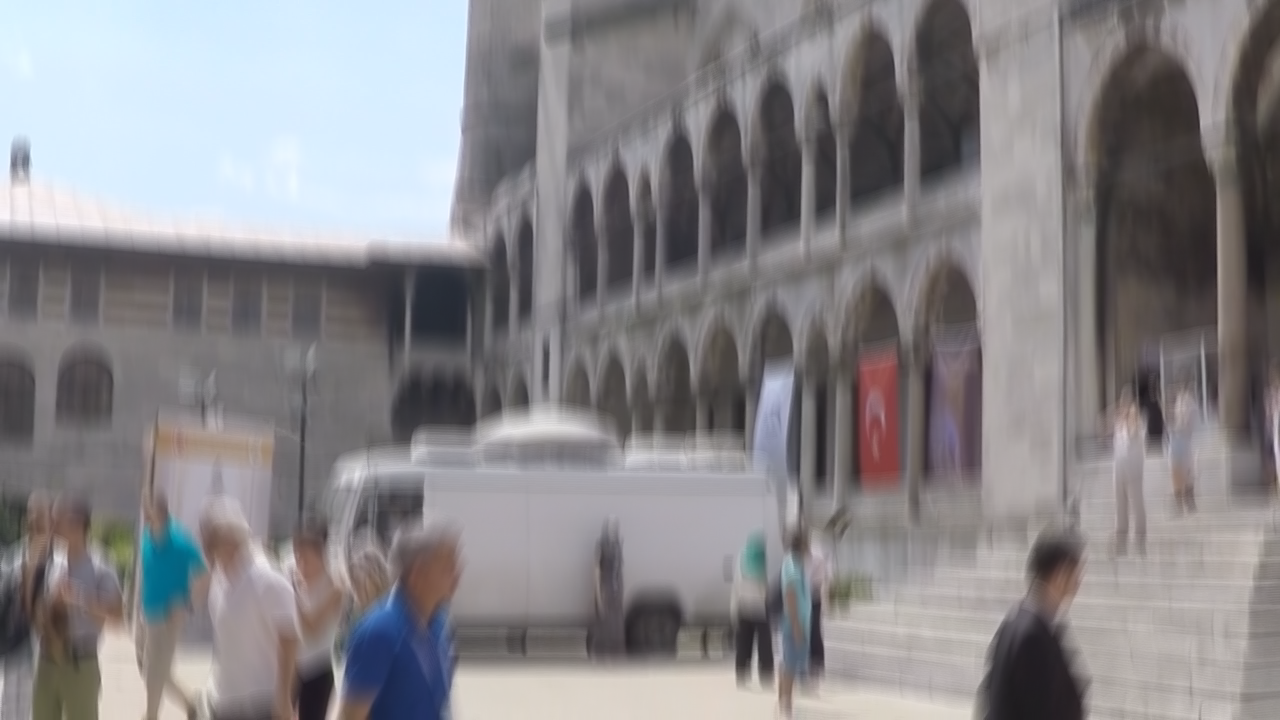}
    \end{subfigure}
    \begin{subfigure}{.2\linewidth}
        \includegraphics[width=\textwidth]{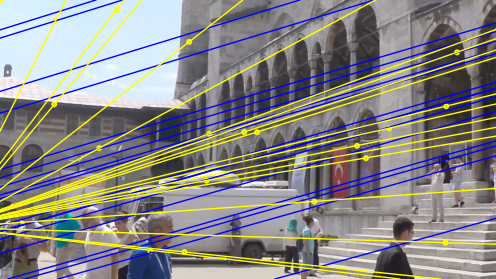}
    \end{subfigure}
    \begin{subfigure}{.2\linewidth}
        \includegraphics[width=\textwidth]{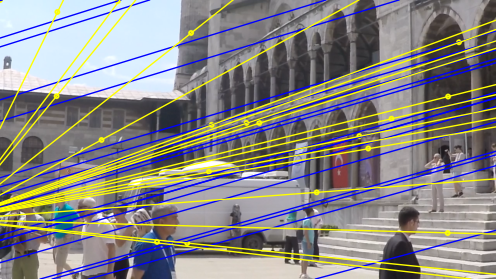}
    \end{subfigure}
    
    \begin{subfigure}{.2\linewidth} 
        \includegraphics[width=\textwidth]{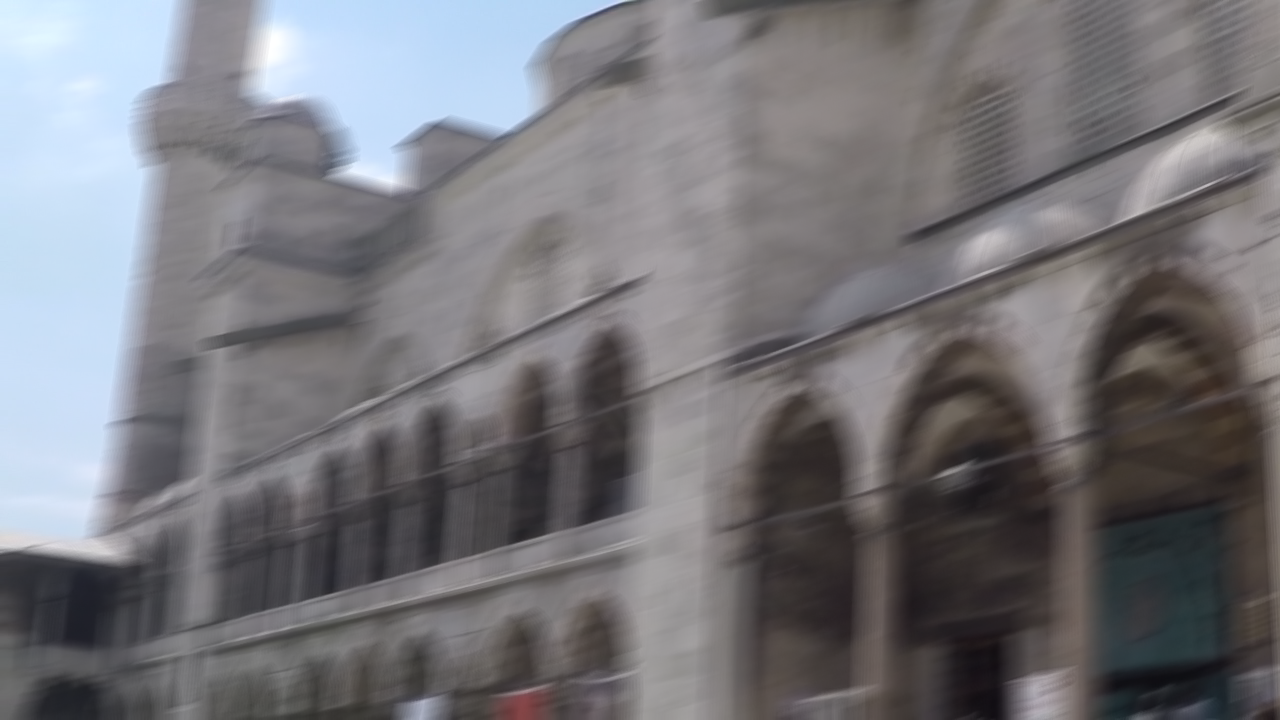}
    \end{subfigure}
    \begin{subfigure}{.2\linewidth}
        \includegraphics[width=\textwidth]{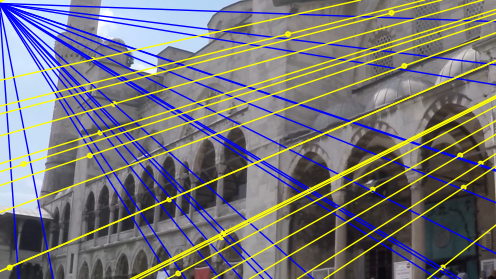}
    \end{subfigure}
    \begin{subfigure}{.2\linewidth}
        \includegraphics[width=\textwidth]{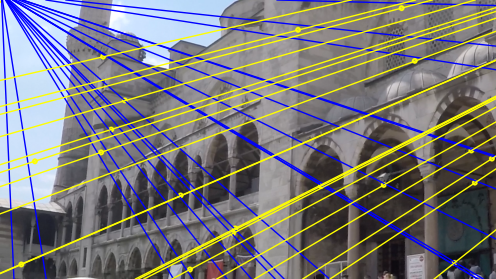}
    \end{subfigure}

    \begin{subfigure}{.2\linewidth} 
        \includegraphics[width=\textwidth]{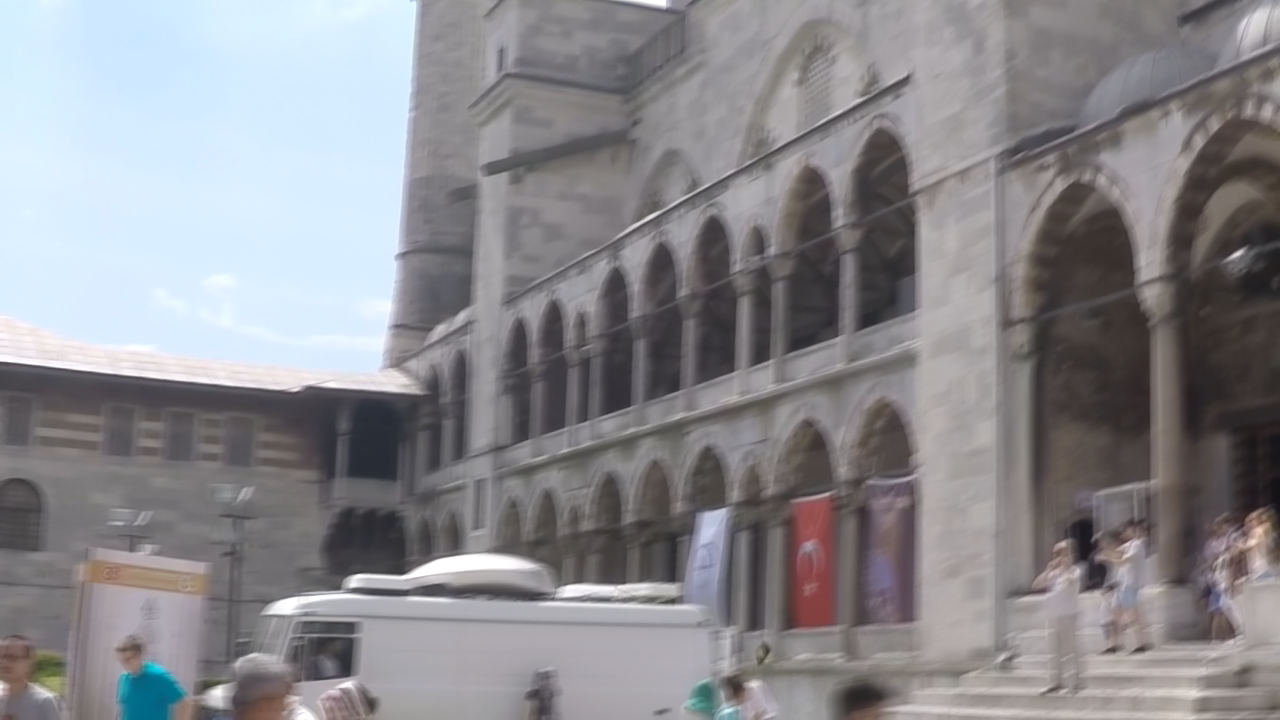}
    \end{subfigure}
    \begin{subfigure}{.2\linewidth}
        \includegraphics[width=\textwidth]{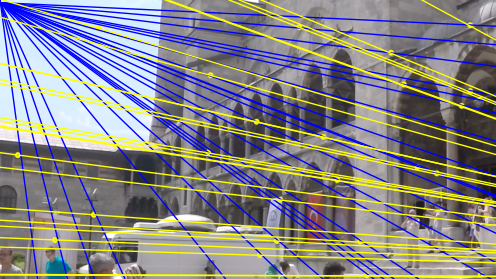}
    \end{subfigure}
    \begin{subfigure}{.2\linewidth}
        \includegraphics[width=\textwidth]{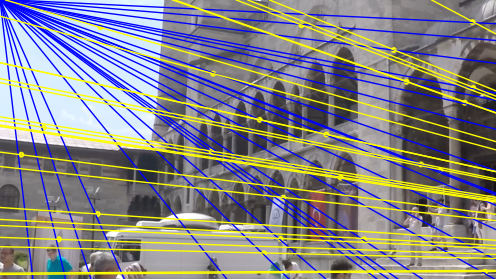}
    \end{subfigure}

    \caption{Examples of problematic cases from the GoPro dataset. $1^{st}$ row: Dynamic scene. $2^{nd}$ row: Dominant plane scene. $3^{rd}$ row: Only camera rotation scene.}
    \label{fig:supp_fmatrix_failure_examples}
\end{figure}

In addition, we also show problematic examples for the fundamental matrix estimation, see Fig.\ \ref{fig:supp_fmatrix_failure_examples}. The $1^{st}$ row depicts a dynamic scene with numerous individual motions present in the blurry input. For such input it is impossible to use a fundamental matrix to summarize the motion from smear paths. 
The $2^{nd}$ and $3^{rd}$ rows showcase the special cases of a dominant plane and a pure rotation scene, respectively. 
In these scenarios, a set of matrices satisfies the epipolar constraint \cite{chum2005degensac}, making it very unlikely that the estimated fundamental matrix  describes the true camera motion. As can be seen in the $2^{nd}$ and $3^{rd}$ rows, although epipolar lines intersect the correspondences, they are very far from pseudo-ground truth (yellow).
Developing methods to detect or properly handle degenerate cases is an interesting direction for future work.

\end{document}